\documentclass[lettersize,journal]{IEEEtran}
\usepackage{amsmath,amsfonts}
\usepackage{algorithmic}
\usepackage{array}
\usepackage[caption=false,font=normalsize,labelfont=sf,textfont=sf]{subfig}
\usepackage{textcomp}
\usepackage{stfloats}
\usepackage{url}
\usepackage{verbatim}
\usepackage{graphicx}
\def\BibTeX{{\rm B\kern-.05em{\sc i\kern-.025em b}\kern-.08em
    T\kern-.1667em\lower.7ex\hbox{E}\kern-.125emX}}
\usepackage{balance}

\usepackage{multirow}
\usepackage{bbding}
\usepackage{amsthm}
\usepackage{amsmath}
\usepackage{threeparttable}
\usepackage{xcolor}

\usepackage[linesnumbered,ruled,vlined]{algorithm2e}

\usepackage{booktabs}
\usepackage[noadjust]{cite}
\usepackage[colorlinks,
            urlcolor=urlcolor,
            linkcolor=mylinkcolor,
            anchorcolor=mylinkcolor,
            citecolor=mylinkcolor]{hyperref}
\usepackage{soul}
\usepackage{color, colortbl}

\definecolor{note_rgb}{rgb}{0,0,1}
\definecolor{mylinkcolor}{RGB}{25, 195, 125}     
\definecolor{urlcolor}{RGB}{0, 131, 208}     

\def\note#1{\textcolor[rgb]{0, 0, 0}{#1}}
\def\noteRA#1{\textcolor[rgb]{0, 0, 0}{#1}}

\newtheorem{example}{Example}

\usepackage{ulem} 

\begin{document}
\title{Towards High-accuracy and Low-latency Spiking Neural Networks with
\note{Two-Stage} Optimization}
\author{Ziming Wang, Yuhao Zhang, Shuang Lian, Xiaoxin Cui,~\IEEEmembership{Member, IEEE}, Rui Yan,~\IEEEmembership{Member, IEEE},\\ and \textsuperscript{*}{Huajin Tang},~\IEEEmembership{Senior Member, IEEE}
\thanks{
    This work was supported by the National Natural Science Foundation of China under Grant 62236007.

    Ziming Wang and Shuang Lian are with the College of Computer Science and Technology, Zhejiang University, Hangzhou 310027, China (e-mail: zi\_ming\_wang@zju.edu.com, slian@zju.edu.cn) 

    Yuhao Zhang is with the Research Center for Intelligent Computing Hardware, Zhejiang Lab, Hangzhou 311100, China (e-mail: zhangyuhao@zhejianglab.com)

    Xiaoxin Cui is with the School of Integrated Circuits, Peking University, Beijing 100871, China (e-mail: cuixx@pku.edu.cn)

    Rui Yan is with the College of Computer Science, Zhejiang University of Technology, Hangzhou 310014, China (e-mail: Ryan@zjut.edu.cn). 
        
    Huajin Tang (Corresponding author) is with the College of Computer Science and Technology and the State Key
    Laboratory of Brain-Machine Intelligence, Zhejiang University,
    Hangzhou 310027, China. 
    (e-mail: htang@zju.edu.cn).
}
}

\markboth{Journal of \LaTeX\ Class Files,~Vol.~18, No.~9, January~2023}%
{How to Use the IEEEtran \LaTeX \ Templates}

\maketitle

\begin{abstract}
    \noteRA{
  Spiking neural networks (SNNs) operating with asynchronous discrete events show higher energy
  efficiency with sparse computation.
    } A popular approach \noteRA{for} implementing deep SNNs is ANN-SNN conversion 
     combining both efficient training of ANNs and efficient inference of SNNs. 
    However, \noteRA{the accuracy loss is usually non-negligible, especially under few time steps, which}
    restricts the applications of SNN on latency-sensitive edge devices greatly. 
    In this paper, we \noteRA{first} identify 
\noteRA{that} such performance degradation stems from the misrepresentation of the negative
    or overflow residual membrane potential in SNNs. Inspired by this, 
    we 
    decompose the conversion error into three \noteRA{parts}: quantization error,
    clipping error, and residual membrane potential
    representation error. With such insights, we
    propose a \note{two-stage} conversion algorithm to minimize
    those errors respectively. 
    \noteRA{{Besides,}} We show each \note{stage} achieves significant performance gains 
    in a complementary manner.
    \note{
    By evaluating on challenging datasets including CIFAR-10, CIFAR-100 and ImageNet, the proposed method demonstrates \noteRA{{the}} state-of-the-art performance in terms of accuracy, latency and energy preservation.}
    Furthermore, 
    \noteRA{our method is evaluated using a more challenging object detection task, revealing}
    notable gains in regression performance under ultra-low latency when compared to existing
    spike-based detection algorithms. Codes are available at https://github.com/Windere/snn-cvt-dual-phase.
\end{abstract}


\section{Introduction}
\IEEEPARstart{S}{pikng} neural networks (SNNs), inspired by mimicking the dynamics of biological neurons \cite{hodgkin1952quantitative,gerstner1993spikes, maass1997networks}, 
have gained increasing interest \cite{roy2019towards}. 
Different from the computation with single, continuous-valued activation
in ANNs, SNNs use binary spikes to transmit and process information.
Each neuron
in the SNNs will remain silent without energy consumption 
until receiving a spike/event afferent \cite{christensen20222022}. 
Such an event-driven computing paradigm enables more power-efficient
solutions on dedicated neuromorphic hardware by 
substituting dense multiplication with sparse addition \cite{han2020rmp}.
As reported in \cite{painkras2013spinnaker,davies2021advancing,merolla2014million}, SNNs on specified neuromorphic processors \note{can} achieve orders of magnitude lower
 energy consumption and latency compared with ANNs.
 In addition, SNNs could synergistically help denoise redundant information
 with inherent temporal dynamics \cite{deng2020rethinking}.
 However, it is still an open problem how to obtain highly-performance SNNs efficiently.

 In general, there are two mainstream methodologies for developing
 \note{supervised deep} SNNs up to date: (1) direct training
 for SNNs \note{and} (2) converting ANNs into SNNs.
 For direct training methods, \noteRA{the back-propagation \noteRA{technique}
 could not be applied to SNNs directly due to 
the threshold-crossing firing in spiking neurons}, which presents challenges for accurately calculating gradients in both spatial and temporal domains.
 Some researchers
 have suggested to circumvent this difficulty by designing a surrogate function and smoothing the non-differentiable
  spike firing \cite{wu2018spatio,shrestha2018slayer,neftci2019surrogate,gu2019stca,fang2021deep}. 
  \noteRA{To efficiently utilize the temporal structure in spike trains, the time-based scheme is proposed to directly assign credits 
  to the shift of spike time in SNNs while taking both inter-neuron and intra-neuron dependencies \cite{bohte2000spikeprop,kim2020unifying,yu2018spike,zhang2020temporal,taherkhani2018supervised,zhang2022rectified} into account.}
Binary probabilistic 
models \cite{pfister2006optimal,gardner2015learning} utilize stochasticity (e.g. the log-likelihood of a spike) to approximate the expectation value of gradients in SNNs. 
Dedicated to improving the performance, transfer learning in SNNs is also explored through transplanting features \cite{ma2022deep} or \noteRA{mitigating} domain shift
\cite{zhan2021effective}.
  Nevertheless,
 it is still hard to train
 large-scale SNNs 
 due to the limited memory capacity,
 the short-time dependency, and the vanishing spike rate in deep
 networks. \noteRA{Moreover, training SNNs on GPUs invariably incurs additional computational and memory overhead, which scales proportionally with the time step \cite{li2021free}, as there is no specific optimization for storage and operations with binary events.}
 
 \noteRA{The alternative approach, known as ANN-SNN conversion, 
involves obtaining
 SNNs from pretrained ANNs,
  which has resulted in some best-performing SNNs on large-scale datasets like
  ImageNet \cite{deng2009imagenet}
 with significantly lower training costs compared to direct training.
 }
 The core idea of conversion is
 to establish the consistent relationship between activations of analog neurons
 and some kind of aggregate representation of spiking neurons, such as spike
 count \cite{wu2021tandem}, spike rate \cite{diehl2015fast,rueckauer2017conversion}, and
  postsynaptic potential (PSP) \cite{deng2021optimal}. 
  With such clear criteria, 
  ANN-SNN conversion has been applied to complex scenarios with competitive performances compared to ANNs 
  \cite{kim2020spiking,tan2021strategy,luo2021siamsnn,kim2022privatesnn}.
Nevertheless, to achieve enough representation precision, 
 considerable simulation steps are usually required for nearly lossless conversion, known as \note{the} accuracy-delay tradeoff.
  It restricts the practical application of SNNs greatly. 
 A large body of recent work \cite{yan2021near, li2021free, bu2021optimal, ho2021tcl} proposes to alleviate this problem 
 by exploiting the quantization and clipping properties of aggregation representations.
 Even though, there is still \note{a significant} performance gap between ANNs and SNNs under low inference latency ($\le$ 16 time steps).
 \noteRA{Some recent works have further bridged this performance gap by introducing \noteRA{a signed neuron model} with a customized spike counter \cite{wang2022signed} or a neuron model firing burst spikes in {a} single time {step}} \cite{yu2021constructing,li2022efficient}. \noteRA{
 However, the underlying cause of performance degradation when mapping ANNs into SNNs is still unclear.} \noteRA{Therefore, our focus is on} analyzing conversion errors and improving the \note{accuracy}-delay tradeoff based on vanilla spiking neurons.
 
 In this paper,
 we \noteRA{explicitly} identify that the conversion error under \noteRA{few time} steps \noteRA{primarily} arises from the misrepresentation of \noteRA{the} residual membrane potential, 
 which \noteRA{{can} accurately
 characterizes} information loss between \noteRA{the} input and output of spiking neurons with asynchronous spike firing.
 Furthermore, we \noteRA{demonstrate that} the error regarding residual potential representation 
 is complementary to quantization and clipping errors.
Inspired by this, we propose an ANN-SNN conversion algorithm with \noteRA{a two-stage optimization approach}
\noteRA{to mitigate these three types of errors}. \noteRA{This approach achieves} remarkable performance with extremely low inference delay.
The main contributions of this work \noteRA{are summarized} as follows:
\begin{enumerate}{}{}
    \item{
        We analyze the operator consistency between ANNs and
    SNNs theoretically and identify the neglected residual
    potential representation problem. Then we divide conversion
    errors into three \noteRA{parts}: quantization error, clipping
    error, and residual potential representation error.
    }
    \item{
     We propose a \note{two-stage} scheme for the
    threefold errors toward lossless conversion under ultra-low inference delay. In the
    first \note{stage}, quantization-clipping functions with trainable
    thresholds and quantization noise are applied to finetune ANNs. 
    In the second \note{stage}, we minimize the
    residual potential error with layer-wise
    calibrations on weights and initial membrane potential. \note{Additionally, we further extend it into ANNs with Leaky ReLU as activation.}
    }
    \item {
    Experimental results on the
    \noteRA{both CIFAR} and ImageNet datasets show
    significant improvements in accuracy-latency tradeoffs 
    compared to state-of-the-art methods \note{across diverse architectures, including ResNet, VGG, ResNeXt, and MobileNet}. 
     For example, we \noteRA{achieved} $70.13\%$ top-1 accuracy ($19.16\%$ improvements)
    on ImageNet with VGG-16 under only 16 time steps.
    }
    \item { A spike-based object detection model is implemented based on the proposed method. The results on the PASCAL VOC dataset demonstrate competitive detection performance 
    with at least 25$\times$ inference speedup
     in \note{comparison} to existing spike-based object detectors. 
     }
\end{enumerate}

\begin{figure}[htb]
    \note{
    \centering
    \includegraphics*[width=.95\linewidth]{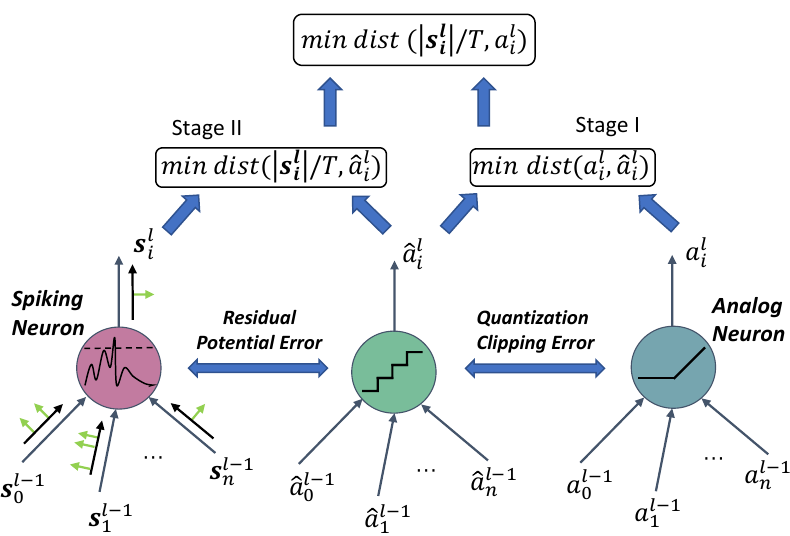 }
    \caption{The criterion of conversion in this article is to minimize the distance of spike rates
    $ r^l_i=| \boldsymbol{s}^l_i|/T$ and ANN activations $a^l_i$ by minimizing three errors with \note{two-stage} methods at each layer.
     Here, $|\boldsymbol{s}^l_i|$ represents the
    cardinal number of spike train $\boldsymbol{s}^l_i$.}
    \label{fig.principle}
    }
\end{figure}
\section{Related Work}


Cao \textit{et al.} \cite{cao2015spiking} first proposed to
convert pretrained ANNs into SNNs and \noteRA{suggested} the criterion of matching ANN activation and spiking rate of SNN.
After the launch of ANN-SNN conversion, the development of conversion algorithms could be divided into two routes in
general. 

\paragraph*{From the perspective of constrained ANNs} Diehl \textit{et al.} \cite{diehl2015fast} \note{found} the importance of weight-threshold balance
and design weight normalization based on the maximum of
layer-wise ANN activations. Afterward, a lot of \note{work was} dedicated to developing more elaborate normalization factors such
as robust normalization \cite{rueckauer2017conversion}, spike-based
normalization \cite{sengupta2019going}, channel-wise normalization \cite{kim2020spiking}, threshold-balancing \cite{xu2017spike}\note{, and} normalization on shortcut connections \cite{hu2021spiking}. 
These methods can be sufficiently 
integrated with ReLU-based ANN to achieve lossless conversion. However, the inference delay is up to hundreds or thousands in general. 
Exploring the characteristics of quantization and clipping in spike rate, 
Yan \textit{et al.} \cite{yan2021near} first
proposed the training scheme with quantization and clipping constraints for ANNs, namely CQ-training. Similarly,
Ding \textit{et al.} \cite{ding2021optimal} proposed the method for weight and threshold training by stages to optimize the upper
 bound of the error between ReLU and CQ activations. Ho \textit{et al.} \cite{ho2021tcl} \note{suggested} a trainable clipping bound 
 in activation functions to balance the threshold. Recently, Bu \textit{et
al.} \cite{bu2021optimal} \note{approximated} the activation of SNNs through a
quantization clip-floor-shift function and \noteRA{explored} the conversion error when the quantization step in ANNs and \noteRA{the} time step
in SNNs are mismatched. In addition to the conversion of full-precision weights, Wang \note{\textit{et al.}} \cite{wang2020deep} further \note{explored} the high-performance conversion under binary weights by rectifying normalization coefficients.

\paragraph*{From the perspective of modified SNNs} The soft reset mechanism is widely adopted \cite{cao2015spiking,han2020rmp}
to avoid information loss from \note{potential} reset. 
Deng \textit{et al.}  \cite{deng2021optimal} decomposed the network conversion error into the layer-wise
conversion error and proposed the extra shift of $\theta/(2T)$ in
spiking neurons to reduce the expectation of quantization error. Comparably, 
Hu \textit{et al.} \cite{hu2021spiking} and Bu \textit{et al.} \cite{bu2022optimized}
configured the initial
 membrane potential of spiking neurons as $\theta/2$ to reduce quantization error.
 Furthermore, Li \textit{et al.}  \cite{li2021free} exploited the calibration
effect of a handful of samples through activation transplanting to reduce clipping error and quantization error.
Yu \textit{et al.}  \cite{yu2021constructing} first show the performance of SNNs could be enhanced greatly with augmented spike and double-threshold \note{schemes}.
Wang \textit{et al.} \cite{wang2022signed} \note{introduced} a signed neuron model with a specific spike counter to 
compensate for the inconsistency between synchronous ANNs
and asynchronous SNNs. Li \textit{et al.} \cite{wang2022signed} \note{enhanced spiking neurons by allowing bursting spikes in a single time step, reducing the information loss.}
In addition to rate coding, customized neuron models \cite{stockl2021optimized, han2020deep} based on temporal coding \note{were} also investigated further to exploit the temporal dynamics of spiking neurons.
 \noteRA{
 The work introduced by Hwang \textit{et al.} \cite{hwang2021lowv} reduced inference latency
by sequentially searching for the optimal initial membrane potential.
 By modulating the 
membrane potential towards a steady firing state of spiking neurons, those methods implicitly alleviate the representation problem from residual potential.
However, without explicitly identifying and formulating
 the error function from residual potential, it is challenging to design targeted optimization strategies.
}


\noteRA{
Different from previous approaches, this work is the first to explicitly identify the source of
error responsible for the stable firing state, called the residual potential error (RPE). Furthermore, we
have introduced an additional fine-tuning stage that is specifically designed
to minimize this particular source of error, thereby enabling high-performance converted SNNs with ultra-low latency.
}

\section{Preliminaries}
\noindent \textbf{a) Analog Neuron Model:} \note{Analog neurons in feedforward
neural networks, such as CNN and MLPs, communicate and
learn with continuous activations.} Mathematically, the forward
computation of the $l$-th layer in feedforward networks
is formed as:
\begin{equation}
    a_{i}^{l}=\sigma\left(z_{i}^{l}\right)=\sigma(\sum_{j} W_{i j}^{l} \cdot a_{j}^{l-1}+b_{i}^{l})
    \label{eq.ann_fw}
    \end{equation}
    where $a^l_i$ is the output of ReLU activation function $\sigma(x) =
    \text{max}(0, x)$. $W^l_{ij}$ and $b^l_i$ are the weights and bias of neuron $i$ in
    the $l$-th layer respectively.

\noindent \textbf{b) Spiking Neuron Model:} 
 Integrate-and-Fire (IF) neuron
model is widely used in conversion algorithms \cite{cao2015spiking, rueckauer2017conversion,sengupta2019going, diehl2015fast,hu2021spiking} because of
the low computing cost and robust representation on firing
rate. At each simulating time step $t$, the IF neuron $i$ receives afferent
spikes $s^{l-1}_i[t]$ and updates its state $u^l_i[t]$ by integrating the
input potential $v^l_i[t]$:
\begin{equation}
    u_{i}^{l}[t]=\hat{u}_{i}^{l}[t-1]+v_{i}^{l}[t]=\hat{u}_{i}^{l}[t-1]+ \sum_{j} W_{i j}^{l}   s_{j}^{l-1}[t]+b_{i}^{l}
    \label{eq.if}
    \end{equation}
where $u^l_i[t]$ and $\hat{u}^l_i[t]$ denote the membrane potential before
and after reset respectively. 
\noteRA{
    The neuron generates a spike $s_{i}^{l}[t]$ and resets
the membrane potential whenever $u^l_i[t]$ exceeds the firing threshold $\theta^l_i$:
}
\begin{figure}[t]
    \centering
    \includegraphics*[width=.99\linewidth]{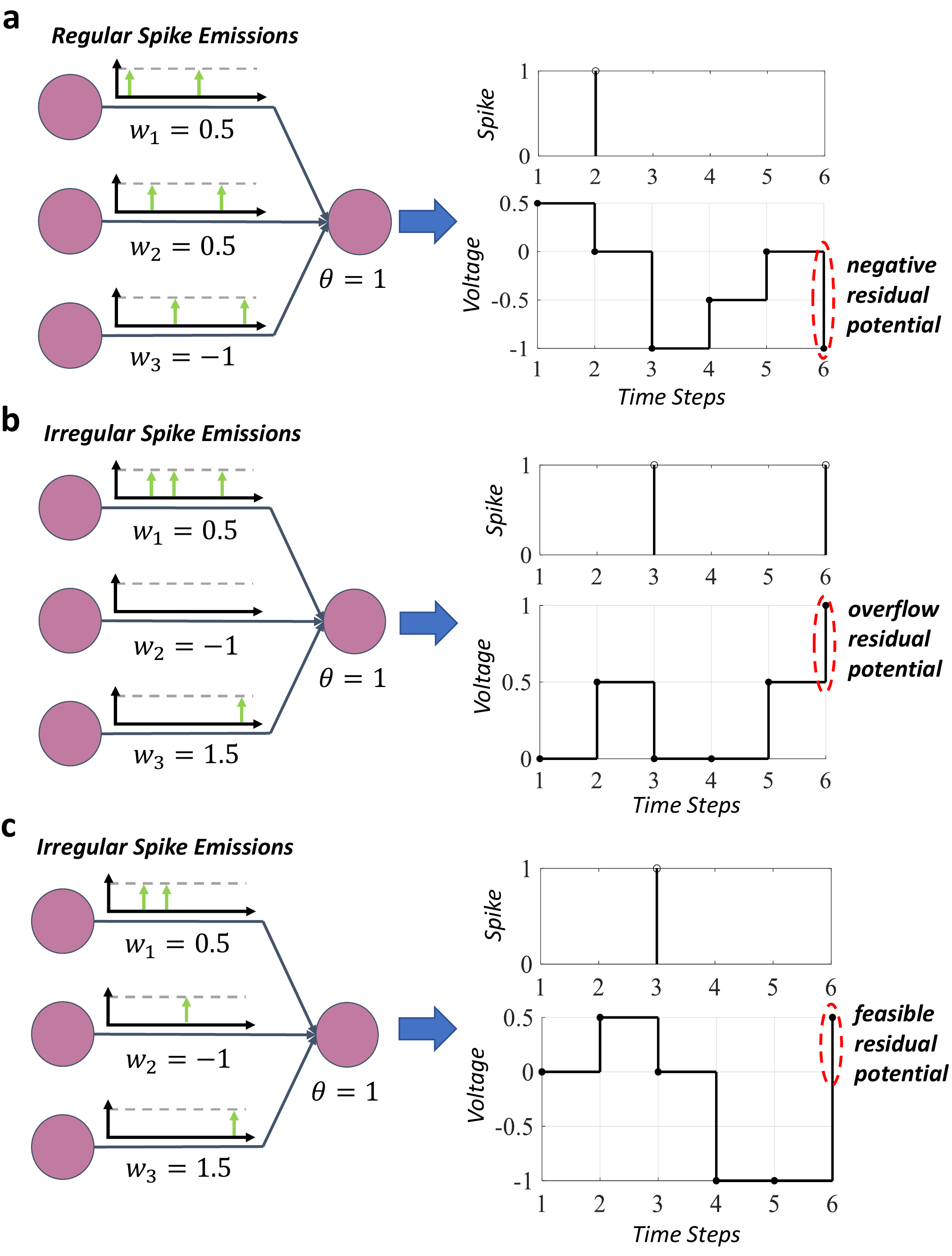}
    \caption{Handcrafted examples under the cases of (a) regular spike emissions with \noteRA{an} undervalued rate, (b) irregular spike emissions with \noteRA{an} overvalued rate, and \note{(c) irregular spike emissions with \noteRA{the} correct rate, to explain the residual membrane potential representation error.}}
    \label{fig.rpe}
\end{figure}
\begin{equation}
    s_{i}^{l}[t]=\Theta\left(u_{i}^{l}[t]-\theta^l_i\right) \text{ with } \Theta(x)= \begin{cases}1, & \text { if } x \geq 0 \\ 0, & \text { otherwise }\end{cases}
    \label{eq.spike}
    \end{equation}
\noteRA{Specifically, we adopt the widely used soft-reset (reset-by-subtraction)
mechanism \cite{han2020rmp,cao2015spiking}
rather than {resetting} as a constant to reduce information loss from the membrane potential over \noteRA{the} threshold.}
 Formally, the soft reset mechanism \noteRA{is} presented as:
\begin{equation}
    \hat{u}_{i}^{l}[t]=u_{i}^{l}[t]-s_{i}^{l}[t] \cdot \theta^{l}_i
    \label{eq.reset}
    \end{equation}
    \noindent \textbf{c) Fusing Batch Normalization:}
    Batch normalization is \noteRA{a crucial component} for most CNN architecture
    as it \noteRA{helps mitigate} the internal covariate shift.
    \noteRA{The absence of batch normalization can be detrimental to network convergence and generalization.}
    However, there is \note{no} \noteRA{direct} equivalent module for batch normalization in SNNs as standard normalization \noteRA{methods damage}
    the binary \noteRA{nature} of spikes. \note{Therefore, as done in most literature \cite{li2021free, deng2021optimal,rueckauer2017conversion, ding2021optimal, yan2021near, hu2021spiking, li2022efficient}, we absorb the batch normalization on activations and \noteRA{transform} it into batch normalization
    on weights and bias at the beginning\noteRA{,} without detrimental effect on performance \cite{jacob2018quantization, nagel2021white}:}
    \begin{equation}
        \begin{gathered}
        \mathrm{BN}_{w}\left({W}_{ij}^{l}\right)=\frac{\gamma_{i}^{l}}{\sqrt{\left(\sigma_{i}^{l}\right)^{2}+\epsilon}} {W}_{ij}^{l} \\
        \mathrm{BN}_{b}\left({b}_{i}^{l}\right)=\frac{\gamma_{i}^{l}}{\sqrt{\left(\sigma_{i}^{l}\right)^{2}+\epsilon}}\left({b}_{i}^{l}-\mu_{i}^{l}\right)+\beta_{i}^{l}
        \end{gathered}
        \end{equation}
     where $\mu^l_i$
     and $\sigma^l_i$
     represent the moving mean and moving variance of ${a}^l_i$ estimated
     from batch statistics during training. $\gamma^l_i$
     and $\beta^l_i$
     are trainable parameters in the batch normalization layer.
     

 \noindent \textbf{d) Input and Readout:} 
Instead of \noteRA{employing} explicit spike coding, we \noteRA{opt to} directly inject the image as input current
 repeatedly into SNNs at each time step, \noteRA{thereby avoiding} information loss as done in \cite{rueckauer2017conversion,wu2019direct}. 
 In \noteRA{essence}, the first layer \noteRA{in} SNNs is responsible for encoding the analog pixels into discrete spikes.
 \noteRA{Additionally},
the average input potential $\overline{v}^l_i = (\sum_{t=1}^T v^l_i[t])/T$ rather than
spike rate $r^l_i= ( \sum_t s^l_i[t])/T$ is read out in the last layer. 
\section{Error Analysis}
\label{sc.ea}
\note{The goal of this paper} is to build a consistent mapping between
spike rates $r^l_i$ in SNNs and activation values $a^l_i$ in
ANNs shown in \note{Fig.} \ref{fig.principle}.  Firstly, we derive the exact relation
between average input potential $\overline{v}^l_i$ and input spike rate $r^{l-1}_j$:
\begin{equation}
    \overline{v}_{i}^{l}=\frac{\sum_{t=1}^{T} \sum_{j} W_{i j}^{l} s_{j}^{l-1}[t]+b_{i}^{l} T}{T}=\sum_{j} W_{i j}^{l} r_{j}^{l-1}+b_{i}^{l}
    \label{eq.r2U}
    \end{equation}
Compared to Eq. \ref{eq.ann_fw}, the functional relation between $\overline{v}^l_i$ and
$r^{l-1}_j$ in IF neurons is identical to that between $z^l_i$ and $a^{l-1}_j$
in analog neurons. Then, the \note{remaining} question is what is the
difference between the function $\overline{v}^l_i \to r^l_i$ and the function
$z^l_i \to a^l_i$? By substituting Eq. \ref{eq.reset} to Eq. \ref{eq.if} and accumulating
from $0$ to $T$, the relation between $r^l_i$ and $r^{l-1}_j$ \noteRA{is} deduced as:
\begin{equation}
    r_{i}^{l}=\sum_{j}\left(W_{i j}^{l} r_{j}^{l-1}+b_{i}^{l}\right) / \theta^l_i-\epsilon_{i}^{l}
    \label{eq.r2r}
    \end{equation}
where $\epsilon^l_i = \frac{\hat{u}^l_i[T]-\hat{u}^l_i[0]}{T\theta^l_i}$ is the component about residual
    membrane potential $\hat{u}^l_i[T]$. By substituting Eq. \ref{eq.r2U} into Eq. \ref{eq.r2r}, 
    we obtain the exact expression between $\overline{v}^l_i$ and $r^l_i$:
    \begin{equation}
        r_{i}^{l}=f_{\theta, T}\left(\overline{v}_{i}^{l}, \epsilon_{i}^{l}\right)=\overline{v}_{i}^{l} / \theta^l_i-\epsilon_{i}^{l}
        \label{eq.sr2sr}
        \end{equation}
Remarkably, $\epsilon^l_i$ is highly nonlinear, nondifferentiable,
and nonconvex as $\hat{u}^l_i[T]$ is calculated with the compound of $T$ Heaviside functions $\Theta(x)$ explicated in \note{Eqs.} \ref{eq.if} - \ref{eq.reset}. Therefore, the only
divergence \noteRA{of} operators appears between the equivalent activation function $f_{\theta,T} (x, \epsilon)$ in SNNs and $\sigma(x)$ in ANNs. To
bypass the non-determinism and complexity of $\epsilon$, we start from the limited residual membrane potential assumption:
\begin{figure}[]
    \centering
    \includegraphics*[width=.8\linewidth]{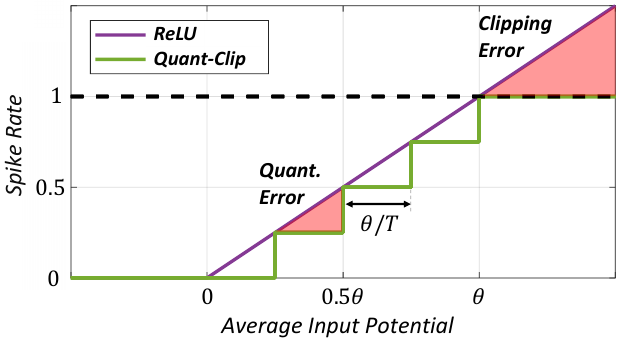}
    \caption{\noteRA{A schematic diagram illustrating quantization and clipping
    errors controlled by threshold $\theta$ and time step $T$ with spike rate representation.}}
    \label{fig.qce}
\end{figure}
\begin{equation}
    0 \leq \hat{u}_{i}^{l}[T]-\hat{u}_{i}^{l}[0]<\theta^l_i
    \label{eq.mem_cond}
    \end{equation}
Under such assumption, $\epsilon^l_i$ \noteRA{is} bounded in $[0, \frac{1}{T})$ whereas $r^l_i \in \{\frac{0}{T},\frac{1}{T},...,\frac{T}{T}\}$.
\note{
The nonlinearity of $f_{\theta,T} (x, \epsilon)$ can be approximated using the floor function with a quantization step of $\theta/T$ 
 (indicated by the green line in Fig. \ref{fig.qce})}:
\begin{equation}
    g_{\theta, T}(x)=\frac{1}{T}\left\lfloor x \frac{T}{\theta}\right\rfloor=\frac{1}{\theta}\lfloor x\rfloor_{\theta / T}
    \label{eq.qce}
    \end{equation}
Furthermore, the spike rate clipping technique is adopted to compensate for the error between $f_{\theta,T}(x)$ and $g_{\theta,T}(x)$
when the condition of Eq. \ref{eq.mem_cond} is not satisfied.
\begin{equation}
    g_{\theta, T}(x)=\operatorname{clip}\left(\frac{1}{\theta}\lfloor x\rfloor_{\theta / T}, 0,1\right)
    \label{eq.qc}
    \end{equation}
However, the errors still exist and remain non-negligible even with such compensation. It is often 
\note{attributed to}
transient dynamics \cite{rueckauer2017conversion}, temporal jitter of spike trains \cite{wu2021tandem}\note{,} or unevenness error \cite{bu2021optimal} without dedicated optimization.
 However, we contend that it is more precise and tractable to attribute such error to \note{an} incomplete representation of residual membrane potential rather than \note{the} irregular
discharge of spikes in IF models.
\begin{figure}[t]
    \centering
    \note{
    \includegraphics[width=.86\linewidth]{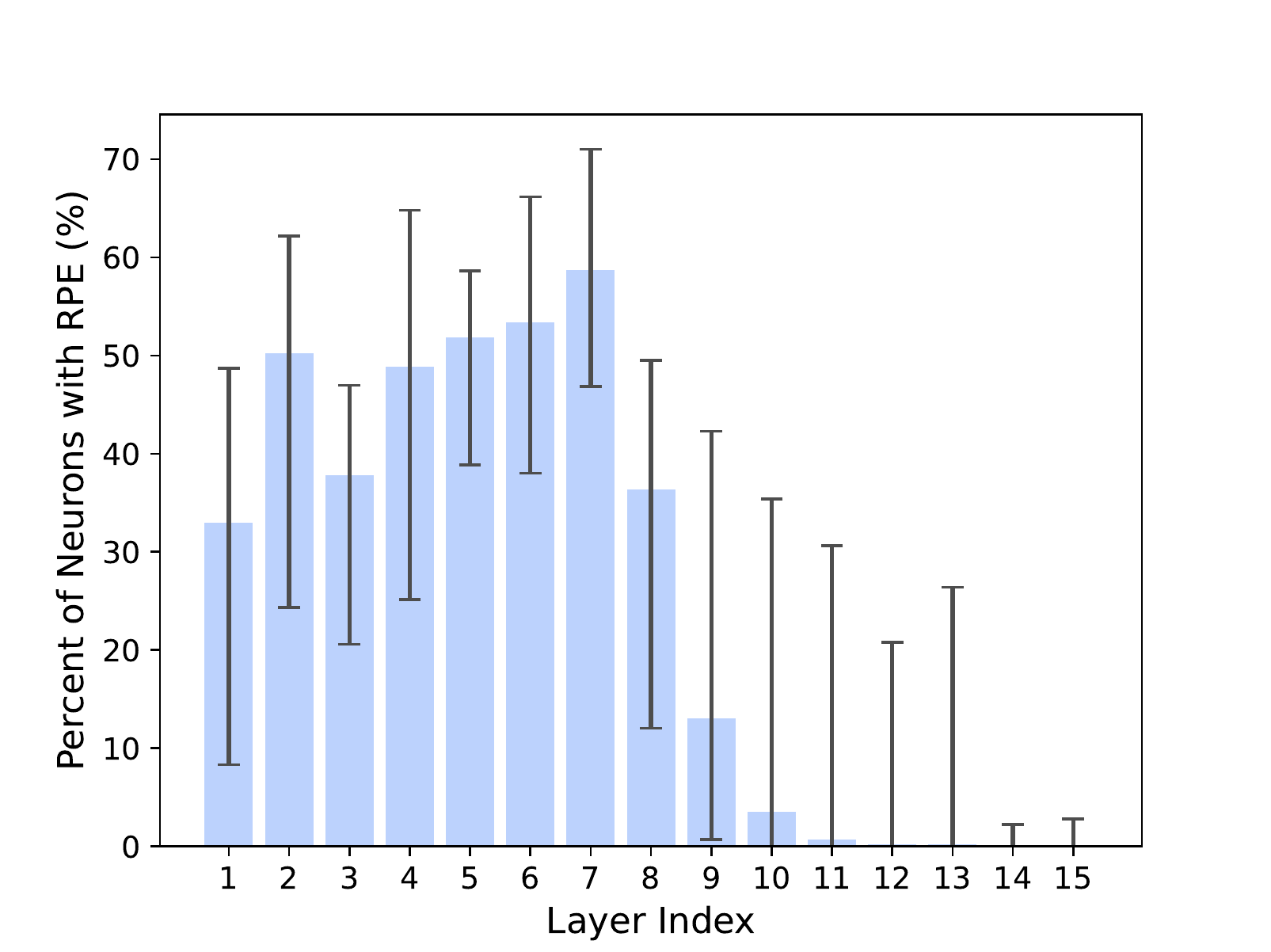 }
    \caption{\noteRA{
        The proportion of neurons with RPE in each layer of converted spiking VGG-16 on the CIFAR-10 dataset. The maximum deviations for all trials are reported, indicating the errors from negative or overflow potential.
    }}
    \label{fig.rpe_per}
    }
  \end{figure}
To \note{illustrate this}, we handcraft \note{three} examples under uniform and irregular \noteRA{spike}
distribution respectively in Fig. \ref{fig.rpe} as follows:
\begin{example}[\note{Regular Spike Emissions with Undervalued Rate}]
    All the input neurons
    emit spikes regularly with an interval of 2 time steps
    in Fig. \ref{fig.rpe}\textbf{a}. However, the output of floor-clipping function
    $g_{\theta,T}(\overline{v})$ is 0 while the real spike rate $r=f_{\theta,T}(\overline{v},\epsilon)$ is $\frac{1}{6}.
$
 The spike rate is undervalued with negative residual potential even under such uniform spike afferents.
\end{example}
\begin{example}[\note{Irregular Spike Emissions with Overvalued Rate}]
    The output neuron
    is \note{in a} saturation state with \noteRA{a} real spike rate $r = \frac{2}{6}$ while the
    expectation $g_{\theta,T}(v)$ is $\frac{3}{6}$ in the example of Fig. \ref{fig.rpe}\textbf{b}. The spike
    rate is overvalued with overflow residual potential.
\end{example}
\begin{example}[\note{Irregular Spike Emissions with Correct Rate}]
    \note{
    The output neuron
    is in a state \noteRA{that yields} the correct output rate $r = g_{\theta,T}(v)= \frac{1}{6}$ as shown in the example of Fig. \ref{fig.rpe}\textbf{c}. The spike
    rate is estimated appropriately with feasible residual potential although the afferent spikes are irregular.
    }
\end{example}
\begin{figure*}[htb]
    \centering
    \note{
    \includegraphics*[width=\linewidth]{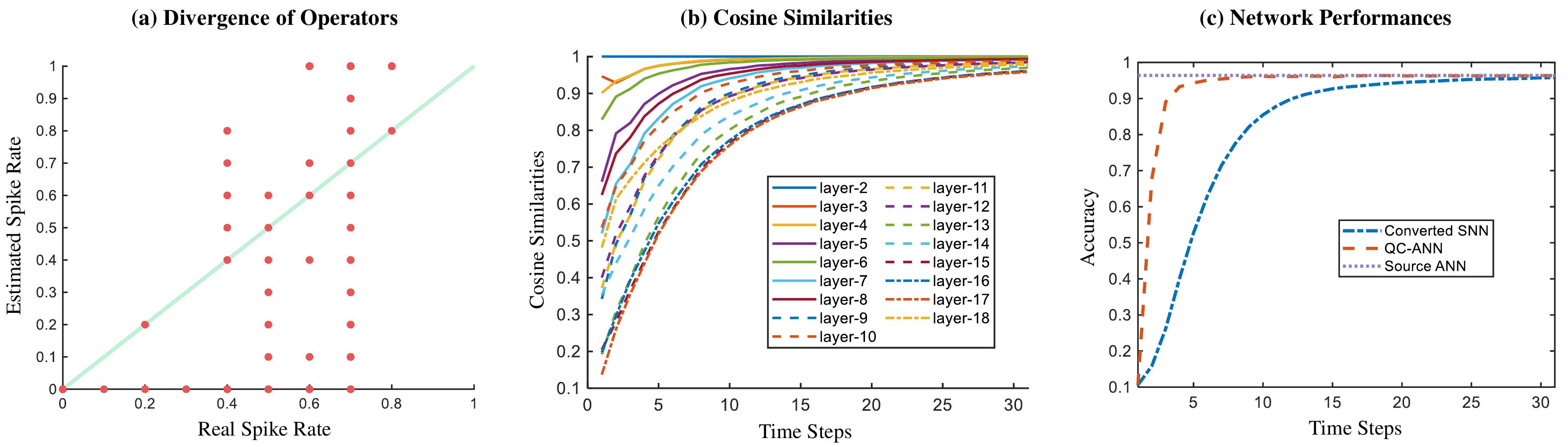 }
    \caption{
            Exploring the influence of RPE over QE and CE.      \textbf{(a)} The non-identical relation between the estimated spike rate and the real spike rate.
       \textbf{(b)}  The \noteRA{layer-wise} cosine similarities between the activation of QC-ANN and the spike rate of converted SNN. 
      \textbf{(c)} The accuracy gap between source ANN, QC-ANN, and converted SNN. }
    \label{fig.prototype}
    }
\end{figure*}
\note{
Therefore, the estimated spike rate may \note{either} surpass or \note{fall} below the \note{actual rate} of IF neurons in both cases as shown in Fig. \ref{fig.rpe}a and Fig. \ref{fig.rpe}b.
\note{In addition, the spike rate can also be evaluated exactly with irregular spike afferents (Fig. \ref{fig.rpe}c).}
 The
fundamental cause of the error \note{is that} the IF neuron with soft-reset mechanism cannot respond to the residual membrane
potential out of $[0, \theta)$.} Then we \note{refer to} the kind of error as residual membrane potential representation error, RPE
in short. It \note{measures} the divergence between the real spike
rate $f_{\theta,T} (x, \epsilon)$ and the estimated spike rate $g_{\theta,T}(x)$ as shown in the left part of Fig. \ref{fig.principle}. Besides,
the quantization and flooring function $g_{\theta,T}(x)$ is divergent from the ReLU function $\sigma(x)$ commonly used by ANNs in
operators. Li \textit{et al.} \cite{li2021free} summarized similar errors in the PSP-based conversion algorithm as clipping error and flooring error.
Here we denote them as clipping error, CE and quantization error, QE generally shown in Fig. \ref{fig.qce}.
\note{Intuitively, RPE reflects the discrepancy that arises from the inherent differences in operating mechanisms between asynchronous SNNs (with temporal unfolding and over-threshold firing) and synchronous ANNs while QE represents the approximating error between binary spikes and real-value activation. 
We conducted an analysis to measure the occurrence of RPE in each layer of the converted spiking VGG16 network using 4 time steps.
Each sample in CIFAR-10 was treated as a separate trial. We recorded the mean and maximum deviation for all trials.
The results, depicted in Fig. \ref{fig.rpe_per}, indicate that RPE occurs frequently in early layers. It is worth noting that RPE at a single layer has a cumulative effect on subsequent layers.
Therefore, RPE is not rare and has a substantial impact on the overall conversion.
 }

In order to explore the influence of RPE over QE and CE further, we adopt the neural network using a quantization-clipping activation function (Eq. \ref{eq.qce}) as an intermediate network called QC-ANN to explicitly separate two kinds of errors.
Specifically, we convert ResNet-18
into its spiking version on the CIFAR-10 dataset and analyze the conversion errors empirically from three perspectives:
\paragraph{\noteRA{Perspective I (Divergence of Operators)}}
We generate 100 periodically injected sequences $\boldsymbol{\mathcal{V}}= \{\boldsymbol{v}_1[t],\boldsymbol{v}_2[t], ... ,
  \boldsymbol{v}_{100}[t]\}$ where each sequence $\boldsymbol{v}[t]$ is sampled with 10 time steps from  
sine functions of different frequencies. By injecting each sequence as current into the single spiking neuron, 
we obtain the real spike rate $\boldsymbol{r}$ 
 through iterative equations (Eq. \ref{eq.if} - Eq. \ref{eq.spike}). Meanwhile, we adopt \noteRA{the} quantization-clipping function $g_{\theta,T}(\frac{1}{T}\sum_t\boldsymbol{v}[t])$ to estimate \noteRA{the} response
spike rate $\boldsymbol{\hat{a}}$.  As shown in Fig. \ref{fig.prototype}\textbf{a}, 
 there is a prominent bias \noteRA{compared to the} identical mapping between $\boldsymbol{\hat{a}}$ and $\boldsymbol{r}$ even in the ideal condition
  which illustrates the fundamental \noteRA{divergence of operators} between 
  $ g_{\theta, T}(\boldsymbol{\overline{v}})$ and $f_{\theta, T}(\boldsymbol{\overline{v}^{l}}, \boldsymbol{\epsilon}^{l})$, i.e. RPE on \note{the} single neuron.
\paragraph{Perspective II  \noteRA{(}Error Accumulation\noteRA{)}} 
We calculate the cosine similarities ($ \operatorname{sim}\left(\boldsymbol{x}_{1}, \boldsymbol{x}_{\mathbf{2}}\right)=\frac{\boldsymbol{x}_{1} \boldsymbol{x}_{2}}{\left\|\boldsymbol{x}_{1}\right\|_{2}\left\|\boldsymbol{x}_{2}\right\|_{2}}$) of activations 
between \noteRA{the} source ANN and QC-ANN on 1024 samples to measure the error accumulation of RPE.
\noteRA{This analysis serves to quantify the accumulation of RPE.}
\noteRA{Upon} comparing the similarity \noteRA{curves} of different layers (see Fig. \ref{fig.prototype}\textbf{b}),
we \noteRA{observe that shallower layers tend to exhibit} higher similarities in general. 
 It demonstrates
that \noteRA{divergence of operations} (RPE) in former layers \noteRA{brings} cumulative errors through subsequent
layers. \noteRA{ Additionally, it is observed that the similarities gradually increase with the growth of the time step $T$. This phenomenon happens because, in this case, the residual term $\boldsymbol{\epsilon}^l = \frac{\hat{\boldsymbol{u}}^{l}[T]-\hat{\boldsymbol{u}}^{l}[0]}{T \theta^{l}}$ diminishes progressively.}
\paragraph{Perspective III \noteRA{(}Performance Gap\noteRA{)}}
To observe the performance gap resulting from cumulative errors,
Fig. \ref{fig.prototype}\textbf{c} displays the accuracy curves of source ANN, QC-ANN, and SNN over various time steps. 
The performance gap between QC-ANN and converted SNN demonstrates the conversion loss caused by neglected RPE about $\boldsymbol{\epsilon}$ while the difference between QC-ANN
and source ANN indicates the impact of QE and CE. 
It is worth noting that the error resulting from RPE generally makes greater contributions 
to the overall conversion error from source ANN to SNN than QE and CE.

The theory and experiments in the analysis above consistently show that RPE works in a complementary manner with respect to QE and CE.
Additionally, as illustrated in Fig. \ref{fig.prototype}\textbf{c}, the neglected RPE gives a wider optimization space for \noteRA{few time} steps. 
Therefore, 
this phenomenon motivates us to 
rethink the conversion errors and redesign the algorithm exploiting the RPE.
\section{Method}
Based on the above error analysis, we propose a \noteRA{two}-stage
conversion scheme (Fig. \ref{fig.workflow}) towards lossless conversion by
minimizing QE, CE, and RPE \note{in} stages. Overall, we \note{start} from a source neural network using \noteRA{the}
 clipping function $\sigma(x)$ as the activation function, then transfer it to a neural
network with trainable quantization-clipping activation function
$g_{\theta,T}(x)$, called QC-ANN later. Therefore, the QE and
CE \noteRA{are} optimized with the fine-tuning of the QC-ANN.
Finally, the layer-wise calibration on weights
and initial membrane potential is adopted to optimize the
RPE based on the activation divergence between $g_{\theta,T}(x)$ and
$f_{\theta,T}(\overline{v}^l_i,\epsilon^l_i)$.
\subsection{\note{Stage-I}: QC-Finetuning with Trainable Threshold}
The principle of the first \note{stage} is to minimize the QE and
CE (right part of Fig. \ref{fig.principle}) and to take the quantization and
clipping property of estimated activation function $g_{\theta,T}(x)$ for
SNNs into ANN training. So we introduce the QC-ANN with
trainable thresholds $\theta^l_i$
as the intermediate network.
\paragraph*{Convert ANNs into QC-ANNs} To avoid the high computational
cost due to training from scratch with QC activation
function $g_{\theta,T}(x)$ under different $T$, we start with training source
ANNs with activation $\sigma_\theta(x) = \text{clip}(\frac{1}{\theta}x, 0, 1)$ unrelated to the specified time
steps $T$. Then we convert the source ANN into QC-ANN by simply copying
the network parameters and replacing the corresponding
activation functions. Here, we adopt \note{a} similar shift term
$\Delta x = \theta/(2T)$ proposed in \cite{deng2021optimal} to minimize
the distance $||\sigma_\theta(x) - g_{\theta,T} (x + \Delta x)||^2$ under different time
steps $T$. So the floor term in $g_{\theta,T} (x)$ in QC-ANN is replaced by a round
term:
\begin{equation}
    \begin{aligned}
    g_{\theta, T}(x) &=\operatorname{clip}\left(\frac{1}{\theta} Q_{\theta / T}(x), 0,1\right) \text { where } \\
    Q_{\Delta}(x) &=\Delta Q_{1}\left(\frac{x}{\Delta}\right)=\Delta \operatorname{round}\left(\frac{x}{\Delta}\right)
    \end{aligned}
    \label{eq.rc}
    \end{equation}
    \paragraph*{Fine-tuning with Quantization Noise} Although the optimal
shift is applied, the QE and CE still exist and contribute
to the conversion error. So we finetune the QC-ANN after
copying weights from source ANN and replacing the activation
function $\sigma_\theta(x)$ with $g_{\theta,T}(x)$. 
Notably, the round term  $Q(x)$ in $g_{\theta,T}(x)$ has null gradients\note{,} which means
the derivative of the input is zero almost everywhere.
To train QC-ANN with such an ill-defined function, we adopt the widely used 
 Straight Through Estimator (STE) \cite{bengio2013estimating} to estimate the gradients:
 \begin{figure}[t]
    \centering
    \includegraphics*[width=\linewidth]{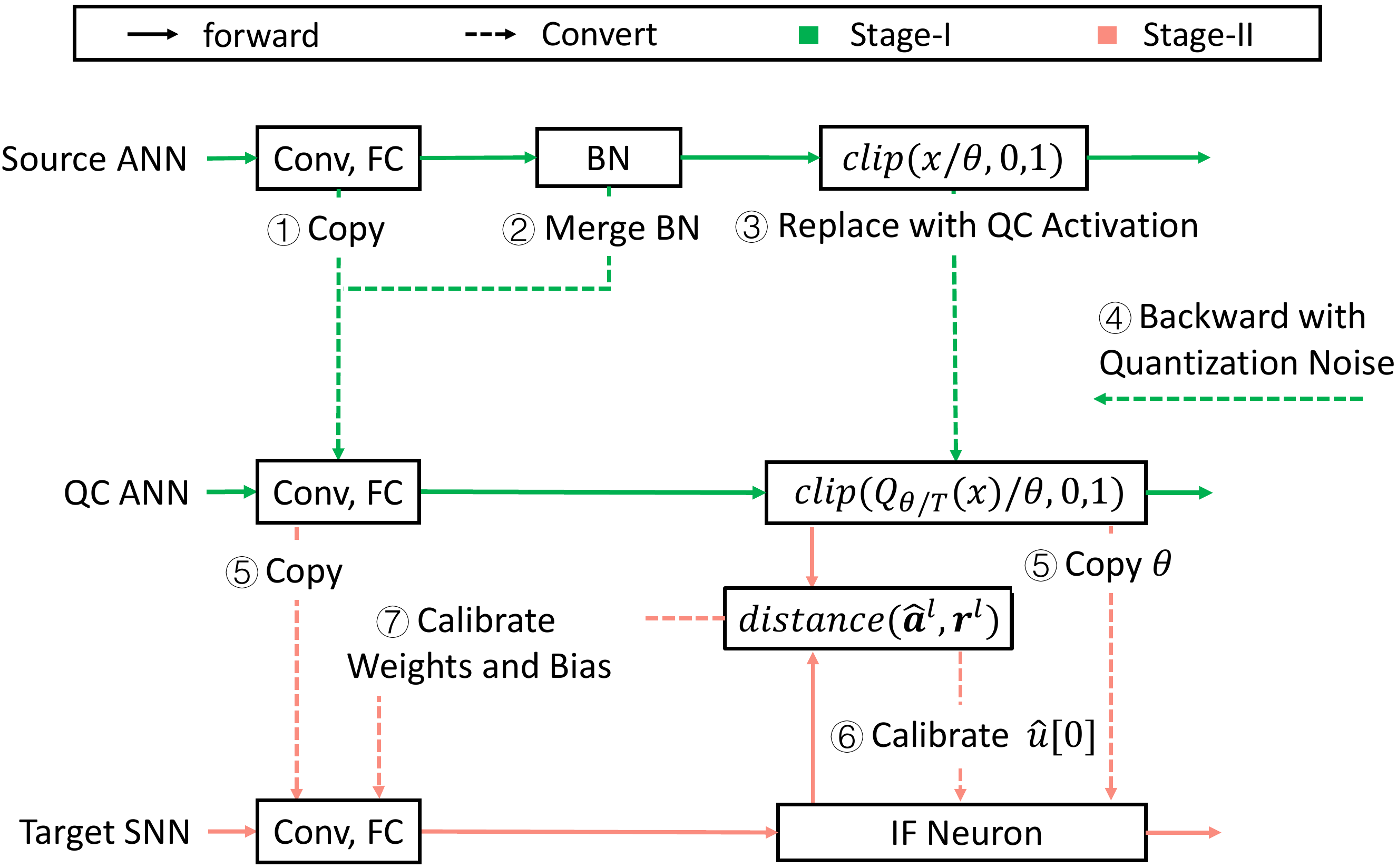}
    \note{
    \caption{\noteRA{The overall conversion framework incorporates a {two-stage} design to minimize threefold errors. The numbers within the circles represent the order of execution.
    }
    \label{fig.workflow}
    }
    }
\end{figure}
\begin{equation}
\frac{\partial Q_{\Delta}(x)}{\partial x}=\frac{\partial Q_{1}(x)}{\partial x}=\frac{\partial\lfloor x\rfloor}{\partial x}=1
\end{equation}
Nevertheless, in the STE scheme, most weights are updated with biased gradients.
 It brings \noteRA{{the}} larger bias with \noteRA{{the}} increasing quantization
step $\theta/T$ when \noteRA{the} time step is relatively low ($T < 8$). Furthermore, the
convergence of \noteRA{the} QC-ANN and the fast-inference ability of converted SNNs are \note{inevitably limited}.
Here we propose to correct the biased gradients using the noisy quantization function 
$\hat{Q}_\Delta(x)$ inspired by \cite{fan2020training}
which introduces Quant-Noise in model compression.
Specifically, we replace a random fraction of $Q_\Delta(x)$ with the
identity mapping $x$ at each forward:
\begin{equation}
    \begin{aligned}
    \hat{Q}_\Delta(x)=x \: + \:  &(Q_\Delta(x)-x) \cdot {M}  \\
    \frac{\partial \hat{Q}_{\Delta}(x)}{\partial x}&=\frac{\partial Q_{\Delta}(x)}{\partial x}=1
    \end{aligned}
\end{equation}
where $ {M}  \sim \operatorname{Bernoulli}(p)$ is the random mask. Then we further modify Eq. \ref{eq.rc} during QC-finetuning as:
\begin{equation}
    g_{\theta, T}(x) =\operatorname{clip}\left(\frac{1}{\theta} \hat{Q}_{\theta / T}(x), 0,1\right) 
 \label{eq.nrc}    
    \end{equation}
Thus, most weights in QC-ANN are updated with unbiased
gradients boosting the convergence under large quantization step.
\subsection{\note{Stage-II}: Layer-wise Calibration with BPTT}
After copying the weights, biases\note{,} and thresholds trained in
QC-ANN into \note{the} target SNN, we get an SNN without considering
RPE. To optimize RPE, the core objective of the second
stage shown in the left part of Fig. \ref{fig.principle}, we adopt a layer-wise calibration
mechanism on weights $\boldsymbol{W}$ and initial membrane potential
$\hat{\boldsymbol{u}}[0]$. 
Firstly, we modify the real activation function $g_{\theta,T} (x)$ on spike rate accordingly with respect to the shift $\theta/(2T)$
adopted in $g_{\theta,T} (x)$:
\begin{equation}
    f_{ \theta,T}(\overline{v}, \epsilon)=\left(\overline{v}+\frac{\theta}{2 T}\right) / \theta-\epsilon
    \label{eq.shift}
    \end{equation}
By calibrating $ f_{ \theta,T}(\overline{v}, \epsilon)$ towards $g_{ \theta,T}(\overline{v})$ further, we \noteRA{effectively narrow} the performance gap without the need for specialized neuron design.
\paragraph*{Coarse Calibration (CC) on Initial Membrane Potential} 

When the
condition Eq. \ref{eq.mem_cond} is not satisfied, the initial membrane potential
$\hat{\boldsymbol{u}}^l[0]$ at the $l$-th layer \noteRA{plays} a coarse adjustment role for
the activation distribution as shown in Eq. \ref{eq.r2r}. Considering the offline strategy of calibration-before-inference on $N$
samples, the core for calibrating $\hat{\boldsymbol{u}}[0]$ lies in solving the
following optimization equation:
\begin{equation}
    \min _{\hat{\boldsymbol{u}}[0]}  \left\{\sum_{i=1}^{N}\left(\frac{\overline{\boldsymbol{v}}_{i}}{\theta}+\frac{1}{2 T}-\frac{\hat{\boldsymbol{u}}_{i}[T]-\hat{\boldsymbol{u}}[0]}{T \theta}-g_{ \theta,T}\left(\boldsymbol{z}_{i}\right)\right)^{2}\right\}
    \label{eq.ccq}
    \end{equation}
    \begin{algorithm}[t]
        \normalem 
        \SetKwData{Left}{left}\SetKwData{This}{this}\SetKwData{Up}{up} \SetKwFunction{Union}{Union}\SetKwFunction{FindCompress}{FindCompress} \SetKwInOut{Input}{Input}\SetKwInOut{Output}{Output}
        \caption{Algorithm for the \note{two-stage} ANN-SNN conversion}
        \label{alg.cvt}
            \Input{ Source ANN model $\mathcal{F}_\text{ANN}(\boldsymbol{W};\boldsymbol{x})$ with activation function $\sigma(x) = \text{clip}(\frac{1}{\theta}x, 0, 1)$; Dataset $D$;  Time step $T$; Total epoch $E$}
            \Output{SNN model $\mathcal{F}_\text{SNN}(\boldsymbol{\hat{W}};\boldsymbol{x})$}
            \BlankLine 
            \noteRA{\tcc{ Stage-I: QC-Finetuning }}
            
            \noteRA{Merge} BN layers into weights and bias.
        
            \For{each layer $\mathcal{F}^l_\text{ANN}(\boldsymbol{W};\boldsymbol{x})$ in $\mathcal{F}_\text{ANN}(\boldsymbol{W};\boldsymbol{x})$}{ 
                \If{$\mathcal{F}^l_\text{ANN}(\boldsymbol{W};\boldsymbol{x})$ is instance of activation function $\sigma_\theta(x)$}
                {
                    replace $\sigma_\theta(x)$ with $ g_{\theta, T}(x)$ (Eq. \ref{eq.nrc}) with the shared parameter $\theta$
                } 
            }
            Finetune the obtained $\mathcal{F}_\text{QC-ANN}(\boldsymbol{W};\boldsymbol{x})$ with $ g_{\theta, T}(x)$ as activation function on $D$
            
            Copy weights from $\mathcal{F}_\text{QC-ANN}(\boldsymbol{W};\boldsymbol{x})$ into the SNN $\mathcal{F}_\text{SNN}(\boldsymbol{W};\boldsymbol{x})$
            and introduce the shift $\theta/2T$ (Eq. \ref{eq.shift})
    
        \noteRA{\tcc{\# Stage-II: Layer-wise Calibration with BPTT }}

            Sample a calibration subset $\hat{D}$ from $D$
    
            \For{$l\leftarrow 1$ \KwTo $\mathcal{F}_\text{ANN}(\boldsymbol{W};\boldsymbol{x})$.layers}
            {
                    Fetch the activation $\hat{\boldsymbol{a}}^l$ of $\mathcal{F}^l_\text{QC-ANN}(\boldsymbol{W};\boldsymbol{x})$ on $\hat{D}$
                    
                    Fetch the output $\boldsymbol{r}^l$ of $\mathcal{F}^l_\text{SNN}(\boldsymbol{W};\boldsymbol{x})$ on $\hat{D}$
    
                    Calibrate initial membrane potential $\hat{\boldsymbol{u}}[0]$ (Eq. \ref{eq.cc})
            }
    
            \For{$l\leftarrow 1$ \KwTo $\mathcal{F}_\text{ANN}(\boldsymbol{W};\boldsymbol{x})$.layers}
            {
                \For{$e \leftarrow 1$ \KwTo E}
                 {
                    \For{$b \leftarrow 1$ \KwTo $\hat{D}$.batches} 
                    {
                        Fetch the activation $\hat{\boldsymbol{a}}^l_b$ of $\mathcal{F}^l_\text{QC-ANN}(\boldsymbol{W};\boldsymbol{x})$ on calibration samples $\hat{D}_b$
                        
                        Fetch the input $\boldsymbol{S}^l=\left[\boldsymbol{s}^l_b[0], \boldsymbol{s}^l_b[1], ..., \boldsymbol{s}^l_b[T] \right]$ of $\mathcal{F}^l_\text{SNN}(\boldsymbol{W};\boldsymbol{x})$ on calibration samples $\hat{D}_b$ 
    
                        Feedforward on the spiking layer $\mathcal{F}^l_\text{SNN}$ wtih $\boldsymbol{S}^l$ and calculate the spike rate $\boldsymbol{r}^l$
    
                        Calculate the gradients $\frac{\partial \mathcal{L}}{\partial \boldsymbol{W}}$ and $\frac{\partial \mathcal{L}}{\partial \boldsymbol{U}[0]}$ (Eq. \ref{eq.bptt})
    
                        Update $\boldsymbol{W}$ and $\boldsymbol{U}[0]$

                }
                }
    
            }
            \Return{$\mathcal{F}_\text{SNN}(\boldsymbol{W};\boldsymbol{x})$}
    
    \end{algorithm}
Suppose $\hat{\boldsymbol{u}}[T]$ is unrelated to $\hat{\boldsymbol{u}}[0]$, then the problem turns into a linear form and the corresponding solution is:
\begin{equation}
    \begin{aligned}
    \hat{\boldsymbol{u}}[0] &=\frac{T \theta}{N} \sum_{i=1}^{N}\left(g_{ \theta,T}\left(\boldsymbol{z}_{i}\right)-f_{\theta,T}\left(\overline{\boldsymbol{v}}_{i} , \epsilon_{i} | \hat{\boldsymbol{u}}[0]=\boldsymbol{0}\right)\right) \\
    &=\frac{T \theta}{N} \sum_{i=1}^{N}\left(\hat{\boldsymbol{a}}_{i}-\boldsymbol{r}_{i}\right)
    \end{aligned}
    \label{eq.cc}
    \end{equation}
By the way, we \note{can} roughly correct the output expectation of the target SNN to match that of the QC-ANN.
Additionally, CC has a relatively low calculation overhead that is proportionate to inference on $N$ calibration samples.


\paragraph*{Fine Calibration (FC) considering Temporal Dynamics}
 To calibrate the RPE
more precisely, we measured the KL divergence between target
activation $\boldsymbol{\hat{a}^l}$ and real spike rate $\boldsymbol{r^l}$ in \note{the} \note{recognition} task:
\begin{equation}
    \min _{\boldsymbol{W}} \mathcal{L}\left(\boldsymbol{r}^{l}, \hat{\boldsymbol{a}}^{l}\right)=D_\text{KL}\left(\hat{\boldsymbol{a}}^{l} \| \boldsymbol{f}_{\theta, T}\left(\overline{\boldsymbol{v}}^{l}(\boldsymbol{W}), \boldsymbol{\epsilon}^{l}(\boldsymbol{W})\right)\right)
    \end{equation}
    Similarly, the MSE loss of Eq. \ref{eq.ccq} is adopted to improve \note{the} numerical precision of coordinate regression in object detection.
    Although $\overline{\boldsymbol{v}}^l$ is the linear combination of $\boldsymbol{W}^l$, $\boldsymbol{\epsilon}^l$ is a superposition
of nondifferentiable Heaviside functions $\Theta(x)$.

In order to 
solve such nonconvex and nondifferentiable optimization
problems, we adopt a rectangular surrogate function $h(u)$ \cite{wu2019direct} to smooth the illness gradient $\frac{\partial \Theta(u)}{\partial u}$:
\begin{equation}
    \frac{\partial \Theta(u)}{\partial u} \approx h(u)=\frac{1}{\alpha} \operatorname{sign}\left(|u-\theta|<\frac{\alpha}{2}\right)
    \end{equation}
    where $\alpha$ controls the smoothness degree for $\frac{\partial \Theta(u)}{\partial u}$.
    Then we \noteRA{obtain} the gradients of weights with backpropagation-through-time (BPTT):
    \begin{equation}
        \begin{aligned}
        \frac{\partial \mathcal{L}}{\partial W_{i j}^{l}}&=\frac{1}{T} \frac{\partial \mathcal{L}}{\partial r_{i}^{l}} \sum_{t^{*}=1}^{T} \sum_{t=1}^{t^{*}} \frac{\partial s_{i}^{l}\left[t^{*}\right]}{\partial u_{i}^{l}[t]} s_{j}^{l-1}[t] & \\
        \frac{\partial \mathcal{L}}{\partial u_{i}^{l}[0]}&=\frac{1}{T} \frac{\partial \mathcal{L}}{\partial r_{i}^{l}} \sum_{t^{*}=1}^{T}  \frac{\partial s_{i}^{l}\left[t^{*}\right]}{\partial u_{i}^{l}[0]} & \\
        \frac{\partial s_{i}^{l}\left[t^{*}\right]}{\partial u_{i}^{l}[t]}&= \begin{cases}h\left(u_{i}^{l}\left[t^{*}\right]\right)  &\text { if } t=t^{*} \\
        \frac{\partial s_{i}^{l}\left[t^{*}\right]}{\partial u_{i}^{l}[t+1]}\left(1-\theta_{i}^{l} h\left(u_{i}^{l}[t]\right)\right)  &\text { if } t<t^{*}
    \end{cases}
        \end{aligned}
        \label{eq.bptt}
        \end{equation}
Compared to the direct training with surrogate gradients using
BPTT, our calibration methods here only feedforward and
backward on one layer which means the spatial complexity is
$O(T \cdot \text{max}_l(n_l))$ rather than $O( T \cdot \sum_l n_l )$ where $n_i$ is the neurons
number at the $l$-th layer in ANNs. Moreover, it is much
easier and more
computationally efficient 
to converge by fine-tuning
on a pretrained layer rather than retraining from scratch on a
large network presented in surrogate gradient methods. Empirically,
it is enough for the calibration with several hundred
of samples. The algorithm is illustrated in \note{the} pseudocode of Algorithm \ref{alg.cvt}.
\section{Experiments}
\begin{table}[]
    \caption{\note{Comparison of data augmentation, encoding, and readout methods used in the proposed approach and other best-performing methods.  The abbreviations RC, RHF, C, AA, CJ, L, and LS stand for RandomCrop, RandomHorizontalFlip, Cutout, AutoAug, ColorJitter, Lighting, and Label Smoothing, respectively.} }
    \label{tb.1t}
    \centering
    \resizebox*{\linewidth}{!}{
        \note{
    \begin{tabular}{|c|c|c|c|c|}
    \hline
    \multirow{2}{*}{\textbf{ Method }} & \multicolumn{2}{c|}{\textbf{Data Augmentation}}                                                              & \multirow{2}{*}{\textbf{Encoding}} & \multirow{2}{*}{\textbf{Readout}} \\ \cline{2-3}
                                       & \textbf{CIFAR}                                      & \textbf{ImageNet}                                                    &                                    &                                   \\ \hline
    { QCFS\cite{bu2021optimal} }      & RC, RHF, C, AA & RC, RF, CJ                 & Direct                             & Analog                            \\
    { RMP \cite{han2020rmp} }          & RC, RHF                & RC, RHF, CJ, L        & Rate                               & -                                 \\
    { OPT \cite{deng2021optimal} }     & RC, RHF, C, AA & RC,RHF,CJ                 & Direct                             & Analog                            \\
    { CAP \cite{li2021free} }          & RC, RHF, C, AA & RC,RHF,CJ                 & Direct                             & Analog                            \\
    { OPI \cite{bu2022optimized} }     & RC,RHF, C, AA & RC,RHF, CJ, LS & Direct                             & Analog                            \\
    RNL \cite{ding2021optimal}                                & RC, RHF                & -                                                           & Direct                             & -                                 \\ \hline
    \textbf{Ours}                      & RC, RHF, C, AA & RC, RHF                             & Direct                             & Analog                            \\ \hline
    \end{tabular}
    }
    }
    \end{table}
\begin{table*}[htb]
    \centering
    \caption{\note{The tradeoff between accuracy and inference delay on the CIFAR-10 and CIFAR-100 datasets. The format "\textbf{A/B}" denotes the reproduced result "\textbf{A}" and the reported result "\textbf{B}" in the original paper, whereas only "\textbf{A}" denotes the reported result.}}
    \label{tb.acc_vs_delay}
    \begin{threeparttable}
    \resizebox{\linewidth}{!}{
    \begin{tabular}{ccccccccccccc}
        \toprule
        
       \textbf{Dataset} & \textbf{ Method } & \textbf{ Architecture }& \textbf{ ANN Acc.}& $T=2$& $T=4$& $T=8$& $T=16$& $T=32$& $T=64$& $T\ge 128$ \\
       \midrule
  &      { TSC \cite{han2020deep}$^\dagger$ } & ResNet-20 &91.47 & - & - & - & - & - & 69.38 & 91.42\\
  \multirow{18}{*}{CIFAR-10} &  { RMP \cite{han2020rmp}$^\dagger$ } & ResNet-20&\note{96.60/91.47} & \note{10.00/-} & \note{10.00/-} & \note{10.30/-} & \note{62.38/-} &\note{91.99/-} & \note{95.62/-} & \note{96.22/91.36}\\

  &      { OPT \cite{deng2021optimal} } & ResNet-20 &\note{96.53/95.46} & \note{10.00/-} & \note{10.02/-} & \note{31.19/-} & \note{81.72/-} & \note{94.21/84.06} & \note{95.97/92.48} & \note{96.38/95.30}\\
  &        { CAP \cite{li2021free} } & ResNet-20 &\note{96.89/95.46} & \note{67.24/-} & \note{80.86/-} & \note{90.44/-}& \note{94.08/-} & \note{95.95/94.78} & \note{96.44/95.30} & \note{96.66/95.45} \\
  &      { RNL \cite{ding2021optimal} } & PreActResNet-18 &\note{93.06/93.06} & \note{10.00/-} & \note{12.19/-} & \note{20.54/-} & \note{43.63/47.63} & \note{82.94/83.95} & \note{91.85/91.96} & \note{93.30/93.41}\\
  &      { OPI \cite{bu2022optimized} } &       ResNet-18                 &\note{96.51/96.04} & \note{34.47/-} & \note{49.94/-} & \note{73.79/75.44} & \note{91.02/90.43} & \note{95.59/94.82} & \note{96.37/95.92} & \note{96.61/96.08}\\
  &      { QCFS \cite{bu2021optimal} } &         ResNet-18                  &96.04 & 75.44 & 90.43 & 94.82 & 95.92 & 96.08 & 96.06 & 96.06\\
    \cmidrule{2-11}
    & \multirow{2}{*}{\textbf{Ours}}   & ResNet-18                 &\textbf{96.41} & \textbf{89.97} & \textbf{93.27} & \textbf{95.36} & \textbf{96.28} & \textbf{96.44} & \textbf{96.49} & \textbf{96.43}\\
  &                                      & ResNet-20                 &\textbf{96.66} & \textbf{90.43} & \textbf{93.27} & \textbf{95.46} & \textbf{96.30} & \textbf{96.60} & \textbf{96.71} & \textbf{96.76}\\
  \cmidrule{2-11}
  &      { TSC \cite{han2020deep} } &  \multirow{7}{*}{VGG-16} &93.63 & - & - & - & - & - & 92.79 & 93.63\\        
  &      { RMP \cite{han2020rmp} } & &\note{95.81/93.63} & \note{10.00/-} & \note{10.00/-} & \note{10.00/-} & \note{26.89/-} & \note{81.13/60.30} & \note{93.34/90.35} & \note{95.45/93.63}\\
  &      { OPT \cite{deng2021optimal} } &                          &\note{95.59/95.72} & \note{10.00/-} & \note{10.00/-} & \note{45.35/-} & \note{91.55/-} & \note{94.66/76.24} & \note{95.43/90.64} & \note{95.53/95.73}\\
  &                  { CAP \cite{li2021free} } &                         &\note{95.75/95.72} & \note{77.78/-} & \note{85.37/-} & \note{90.82/-} & \note{93.41/-} & \note{94.72/93.71} & \note{95.35/95.14} & \note{95.62/95.79}\\
  &                { RNL \cite{ding2021optimal} } &                          &\note{92.81/92.82} & \note{10.00/-} & \note{10.00/-} & \note{11.17/-} & \note{52.91/57.90} & \note{85.00/85.40} & \note{91.04/91.15} & \note{92.52/92.51}\\
  &                { OPI \cite{bu2022optimized} } &                          &\note{95.59/94.57} & \note{68.37/-} & \note{88.03/-} & \note{93.05/90.96} & \note{94.74/93.38} & \note{95.42/94.20} & \note{95.51/94.45} & \note{95.55/94.55}\\
  &                { QCFS \cite{bu2021optimal} } &                          &95.52  & 91.18 & 93.96 & 94.95&95.40 & 95.54 & 95.55 &95.59\\
  \cmidrule{2-11}
  &  \textbf{Ours}                    &       VGG-16                  &\textbf{95.73} & \textbf{91.71} & \textbf{94.06} & \textbf{95.26} & \textbf{95.55} & \textbf{95.71} & \textbf{95.78} & \textbf{95.81}\\
       \midrule \multirow{14
       }{*}{CIFAR-100}
        &   { RMP \cite{han2020rmp} } & \multirow{5}{*}{VGG-16}  &71.22 & \note{1.00/-} & \note{1.00/-} & \note{1.00/-} & \note{8.47/-} &\note{39.04/-} & \note{65.09/-} & \note{74.47/63.76}\\
  &        { OPT \cite{deng2021optimal} } &  &\note{77.09/\textbf{77.89}} & \note{1.00/-} & \note{1.00/-} & \note{2.32/-} & \note{53.37/-} & \note{71.80/7.64} & \note{76.13/21.84} & \note{77.02/55.04}\\
  &        { CAP \cite{li2021free} } & &\note{77.54/\textbf{77.89}} & \note{30.82/-} & \note{44.16/-} & \note{61.27/-} & \note{68.27/-} & \note{72.41/73.55} & \note{76.34/76.64} & \note{77.33/\textbf{77.40}}\\
  &        { OPI \cite{bu2022optimized} } &         &            \note{77.09/76.31} & \note{38.42/-} & \note{51.67/-} & \note{64.63/60.49} & \note{72.72/70.72} & \note{76.15/74.82} & \note{76.91/75.97} & \note{77.13/76.31}\\
  &        { QCFS \cite{bu2021optimal} } &             &            76.28 & 63.79 & 69.62 & 73.96 & 76.24 & 77.01 & 77.10 & 77.08\\
  \cmidrule{2-11}

          &  \textbf{Ours}          &  VGG-16                        &77.22 & \textbf{64.89} & \textbf{70.08} & \textbf{75.06} & \textbf{76.61} & \textbf{77.41} & \textbf{77.35} & 77.24 \\
          \cmidrule{2-11}
          &  { TSC \cite{han2020deep}$^\dagger$ } & \multirow{4}{*}{ResNet-20} &68.72 & - & - & - & - & - & - & 58.42\\
  &         { RMP \cite{han2020rmp}$^\dagger$} &   &\note{80.99/68.72} & \note{1.00/-} & \note{1.00/-} & \note{1.03/-} & \note{3.99/-} &\note{51.40/27.64} & \note{75.25/46.91} & \note{79.63/57.69}\\
  &        { OPT \cite{deng2021optimal} } & &\note{79.20/77.16} & \note{1.00/-} & \note{1.02/-} & \note{1.65/-} & \note{34.57/-} & \note{71.31/51.27} & \note{78.07/70.12} & \note{79.07/75.81}\\
  &        { CAP \cite{li2021free} } &  &\note{80.38/77.16} & \note{32.98/-} & \note{51.44/-} & \note{67.85/-} & \note{75.27/-} & \note{78.80/76.32} & \note{80.04/77.29} & \note{80.51/77.73}\\
  &        { OPI \cite{bu2022optimized} } &         ResNet-18              &\note{79.43/79.36} & \note{33.07/-} & \note{47.16/-} & \note{65.50/57.70} & \note{75.37/72.85} & \note{78.60/77.86} & \note{79.29/78.98} & \note{79.60/79.60}\\
  &        { QCFS \cite{bu2021optimal} } &          ResNet-18                &78.80 & 70.79 & 75.67 & 78.48 & 79.48 & 79.62 & 79.54 & 79.61 \\
  \cmidrule{2-11}
  &  \multirow{1}{*}{\textbf{Ours}}          &  ResNet-18                      &{79.22} & \textbf{73.20} & \textbf{75.98} & \textbf{78.84} & \textbf{79.83} & \textbf{80.36} & \textbf{80.61} & \textbf{80.41} \\
        \bottomrule
    \end{tabular}
 }
    \begin{tablenotes}
    \item The symbol $\dagger$ indicates using standard ResNet-20 architecture 
    while others construct ResNet-20 by adding two more layers to the ResNet-18. 
    \end{tablenotes}

    \end{threeparttable}

    \end{table*}
In this section, we first
 evaluate the effectiveness and efficiency
of the proposed method compared to other \noteRA{state-of-the-art} 
conversion algorithms for image \note{recognition} tasks on CIFAR-10, CIFAR-100\note{,} and ImageNet datasets.
Each component of the design is \note{validated} with \note{a} \noteRA{comprehensive} ablation study.
Subsequently, both sample efficiency and energy efficiency are \note{examined} in different time steps.
Furthermore, we investigate spike-based object detection with the \noteRA{proposed} \note{two-stage} pipeline in \note{shortened} time steps.


\subsection{Implementation Details}
\paragraph{Training of Source ANN} To obtain a high-performance source ANN trained with
$\sigma_\theta(x) = \text{clip}(\frac{1}{\theta}x, 0, 1)$. In general, there are three approaches:
(1) Training an ANN with clipping function $\sigma_\theta(x)$ directly as done in \cite{deng2021optimal}. (2)
training an ANN with ReLU function and then constrain the activations into $[0,1]$ with weight-threshold balancing technique
\cite{diehl2015fast,rueckauer2017conversion,kim2020spiking}. 
(3) Training an ANN with the rescaled clipping function $\sigma_\theta(x) = \text{clip}(x, 0, \theta)$ \cite{ho2021tcl} and then fusing the scale factor $\theta$ 
into weights after training. 
In experiments, we adopt the first method for CIFAR-10. For CIFAR-100 and ImageNet,
we use the third method to achieve higher performance.
\paragraph{Experiments Settings}
For CIFAR-10 and CIFAR-100 \note{datasets}, we crop images into 32$\times$32 with the padding of 4 pixels and flip them horizontally at random.
Besides, as done in \cite{li2021free,bu2021optimal}, we adopt the data augmentation of Cutout \cite{devries2017improved} and AutoAugment \cite{cubuk2019autoaugment}
to increase the generalization ability of source ANNs. 
For ImageNet, we adopt the standard pre-processing pipeline and crop images into 224$\times$224. 

For training source ANN,
we initialize the learning rate as 0.1 and then update it through
the SGD optimizer with \note{a} momentum of 0.9. 
Meanwhile, the weight decay is set to 5e-4 to regularize the
network. Besides, the learning rate is set as 1e-4 further to finetune QC-ANN.
For CIFAR-10, the probability of quantization noise $p$ is
set as 0.1 and 0.2 when $T \le 4$. Otherwise, no noise \noteRA{is}
injected into QC-ANNs. The $p$ is set as 0.1 across all time steps on ImageNet and CIFAR-100 datasets. 
\note{For \note{Stage II}, we observed that the errors tend to vary across different layers. Therefore, to address this issue and facilitate layer-wise calibration with BPTT presented in Stage-II, we opted to use the Adam optimizer with an adaptive learning rate.  The initial learning rate is configured as 5e-4 and 1e-4 for recognition and detection tasks, respectively.} The weight decay is set as 0.
Moreover, we also adopt layer-wise BPTT calibration on the initial potential for the ImageNet dataset. 
For energy analysis, we sample
1024 samples and calculate the mean value and standard deviation of synaptic operations.
Early stopping is also used \note{to} alleviate overfitting with \note{a} tolerance of 20 epochs.
\paragraph{Evaluation Metrics} For the task of object detection, we report the mean Average Precision (mAP) with different localization requirements.
As an illustration, AP@0.75 denotes the mAP score which considers the predictions with a correct classification and localization $\text{IoU} \ge 0.75$ as the true positive.
Furthermore, according to the size of \note{the ground} truth, the mAP \noteRA{is} divided into that of small objects AP$_S$, middle objects AP$_M$, and large objects AP$_L$. 
\begin{figure}[t]
    \centering
\resizebox*{\linewidth}{!}{
    \includegraphics*[width=\linewidth]{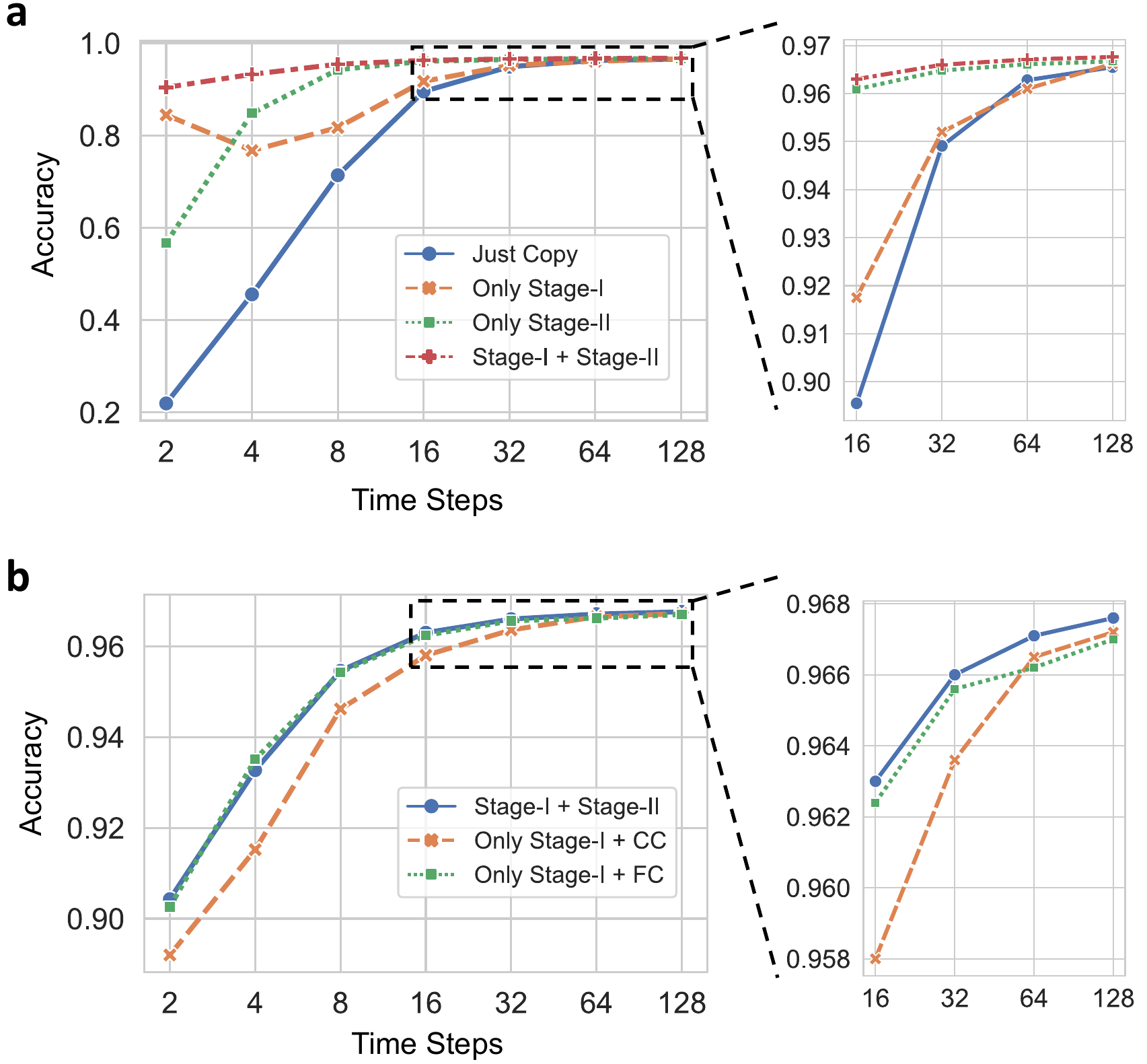}
}

    \caption{ \noteRA{
        Exploring the impact of individual components. (\textbf{a}) Each stage contributes to enhancing the performance in a complementary way. (\textbf{b}) The ablation study of the two calibration modules within Stage II.
    } }
    \label{fig.ablation}
\end{figure}
\subsection{\note{Ablation} Study}
In this section, we conduct a series of ablation studies and show the proposed method 
reduces the conversion loss in a complementary manner. Specifically, we test ResNet-20 on
CIFAR-10 from $T = 2$ to $T = 128$ 
under six conditions: \textit{Just Copy Weights}, \note{\textit{Only Stage-I}}, \note{\textit{Only Stage-II}}, \note{\textit{Stage-I}
+ \textit{Stage-II}}, \note{\textit{Stage-I + CC}}, and \note{\textit{{Stage-I + FC}}}.

As shown in Fig. \ref{fig.ablation}\textbf{a},
both \note{\textit{Only Stage-I}} and \note{\textit{Only Stage-II}}
 improve the performance effectively, which is especially
impressive when $T\le32$.
Besides, \note{\textit{Stage-I + Stage-II}} with dedicated optimization
obtains the best results across all time steps which indicates the synergic effect of the two \note{stages}.
 Comparing the two \note{stages}
individually, we find \note{\textit{Stage-I}} optimizing the QE and CE 
\noteRA{achieves} better results when $T < 4$. Otherwise, \note{\textit{Stage-II}} optimizing
RPE mainly \noteRA{obtains} higher performance. It
is interpretable as 
\noteRA{the quantization factor $\theta/T$ and the QE naturally grow as $T$ decreases.}
So the performance
degradation stemmed from the QE plays a
dominant role compared to RPE when $T$ is extremely low. 
\noteRA{
Surprisingly, {\textit{Only Stage-II}} only needs 8 time steps to match the performance of
\textit{Stage-I + Stage-II}, whereas \textit{Only {Stage}-I} requires 32 time steps.
}
It demonstrates that optimizing RPE is more efficient in low-latency conversion in general. 
 So simply adopting layer-wise \note{Stage-II} with a handful of samples is
enough for conversion on resource-constrained devices. Notably,
there is a strange accuracy drop for \textit{Only \note{Stage}-I} when
$T = 4$. It may come from increasing RPE which is not optimized
by \note{Stage-I}.  As shown in the curve of \textit{\note{Stage-I + {Stage}-II}}, the drop 
is eliminated by introducing \note{Stage-II} (minimizing RPE). This indeed supports the 
efficacy of minimizing RPE.

To understand the effect of each component in \note{Stage-II} further, we do
coarse calibration and fine calibration separately after \note{Stage-I} (Fig. \ref{fig.ablation}\textbf{b}).
In general, both components improve performance consistently. In particular, 
we find \textit{\note{Stage-I +  FC}} \noteRA{achieves} extremely close results to 
\textit{\note{Stage-I + Stage-II}}. And it even outperforms \textit{\note{Stage-I + Stage-II}} when $T=4$.
Nevertheless, coarse calibration still provides a 
 lightweight solution without gradient computation and a proper initialization for
fine calibration.

\begin{table}[t]
    \centering
    \caption{\note{{Comparing relative performance loss under $T=32$, $64$, and $128$ on ImageNet Dataset.  }} }
    \label{tb.relative}
    \note{
      \begin{threeparttable}
    \resizebox{ \linewidth}{!}{
      \begin{tabular}{ccccc}
        \toprule
        \multirow{2}{*}{\textbf{ Method }}        & \multirow{2}{*}{\textbf{ Architecture }} & \multicolumn{3}{c}{$(\text{Acc}_{\text{ANN}}-\text{Acc}_{\text{SNN}})/\text{Acc}_{\text{ANN}}$}                        \\ \cmidrule{3-5} 
                                                  &                                          & $T=32$ & $T=64$             & $T=128$                     \\ \hline
        { QCFS \cite{bu2021optimal} }             &  \multirow{6}{*}{ResNet-34}                                                 &   6.66\%       & 2.65\%              & 1.57\%                       \\
        { RMP \cite{han2020rmp} }                 &                        &  98.35\%       & 56.91\%     & 15.09\%             \\
        { Norm-on-Shortcut \cite{hu2021spiking} } &                                          &  89.74\%        &  34.71\%     &  8.04\%               \\
        { TCL \cite{ho2021tcl} }                  &                                          &      98.64\%   & 72.82\%     &   14.20\%            \\
        { OPT \cite{deng2021optimal} }            &                                          &       93.61\% &  35.69\% & 7.08\%          \\
        { CAP \cite{li2021free} }                 &                                          &       10.40\% & 3.89\% & 1.65\% \\ \midrule
        \textbf{Ours}                             & ResNet-34                                &  \textbf{2.41\%}      & \textbf{1.06\%}     &\textbf{ 0.08\% }                    \\ \midrule
        { RMP \cite{han2020rmp} }                 & \multirow{7}{*}{VGG-16}                  &    99.57\%    & 87.21\%      & 31.72\%              \\
        { OPT \cite{deng2021optimal} }            &                                          &     98.53\%   & 69.62\% & 13.58\%          \\
        { CAP \cite{li2021free} }                 &                                          &    11.45\%   & 5.64\% & 2.82\%          \\
        SNM \cite{wang2022signed}                 &                                          &        11.48\%   & 2.30\% & 0.44\%                    \\
        Burst \cite{li2022efficient} $\dagger$              &                                          &         4.93\%   & 1.28\% & 0.38\%                      \\
        { QCFS \cite{bu2021optimal} }             &                                          &     7.83\%   & 1.94\% & 0.43\%                              \\
        { OPI \cite{bu2022optimized} }            &                                          &     13.56\%   & 3.18\% & 0.81\%                     \\ \midrule
        \textbf{Ours}                             & VGG-16                                   &   \textbf{1.92\%}     & \textbf{0.27\%}    & \textbf{-0.07\%}              \\ \bottomrule
        \end{tabular}
    }
      \begin{tablenotes}
          \item $\dagger$ denotes the method adopts multi-bit outputs rather than binary outputs.
      \end{tablenotes}
      \end{threeparttable}
    }
  \end{table}

\begin{table*}[htb]
    \centering
    \caption{The tradeoff between accuracy and inference delay on ImageNet Dataset. \note{The format "\textbf{A/B}" denotes the reproduced result "\textbf{A}" and the reported result "\textbf{B}" in the original paper, whereas only "\textbf{A}" denotes the reported result.}
    }
    \label{tb.imagenet}
    \resizebox{ 0.95\linewidth}{!}{
    \begin{tabular}{cccccccccc}
    \toprule \textbf{ Method } & \textbf{ Architecture } & \textbf{ ANN Acc.}& $T=4$& $T=8$& $T=16$& $T=32$& $T=64$& $T=128$& \note{$T= 256$} \\
    \midrule   
    { TSC \cite{han2020deep} } &       \multirow{7}{*}{ResNet-34}                  &70.64  & - & - & - & - & - & - & \note{55.65}\\
    { QCFS \cite{bu2021optimal} } &                        &74.32 & - & - & 59.35&69.37 & 72.35 & 73.15 &\note{73.37}\\

    { RMP \cite{han2020rmp} } &  & \note{74.54/70.64}& \note{0.10/-} & \note{0.10/-} & \note{0.10/-} & \note{1.23/-} & \note{32.12/-} & \note{63.29/-} & \note{71.20/55.65}\\
    { Norm-on-Shortcut \cite{hu2021spiking} } & & \note{74.54/72.88}  & \note{0.13/-} & \note{0.20/-} & \note{0.70/-} & \note{7.65/-} & \note{48.67/-} & \note{68.55/-} & \note{72.75/-}\\
              { TCL \cite{ho2021tcl} } &                         &\note{73.43/70.85} & \note{0.09/-} & \note{0.10/-} & \note{0.18/-} & \note{1.00/-} & \note{19.96/-} & \note{63.00/-} & \note{72.57/-} \\
                { OPT \cite{deng2021optimal} } &                         &\note{73.43/\textbf{75.66}} & \note{0.10/-} & \note{0.10/-} & \note{0.15/-} & \note{4.69/0.09} & \note{47.22/0.12} & \note{68.23/3.19} & \note{72.37/47.11
                }\\
              { CAP \cite{li2021free} } &                         &\note{75.65/\textbf{75.66}} & \note{1.68/-} & \note{23.82/-} & \note{52.72/-} & \note{67.78/64.54} & \note{72.71/71.12} & \note{74.40/\textbf{73.45}} & \note{74.92/74.61}\\

              \midrule \textbf{Ours}                    &       ResNet-34 & {73.43} & \textbf{55.71} & \textbf{61.20} & \textbf{67.77} & \textbf{71.66} & \textbf{72.65} & {73.37} & 73.45\\
\midrule  
{ TSC \cite{han2020deep} } &        \multirow{8}{*}{VGG-16}                   &73.49 & - & - & - & - & - & - & 69.71\\
{ RMP \cite{han2020rmp} } && \note{74.53/73.49} & \note{0.08/-} & \note{0.09/-} & \note{0.13/-} & \note{0.32/-} & \note{9.53/-} & \note{50.89/-} & \note{68.78/48.32}\\

                { OPT \cite{deng2021optimal} } &                         &\note{74.88/\textbf{75.36}} & \note{0.10/-} & \note{0.11/-} & \note{0.21/-} & \note{1.10/0.114} & \note{22.75/0.118} & \note{64.71/0.122} & \note{72.81/1.81}\\
                { CAP \cite{li2021free} } &                         &\note{74.53/\textbf{75.36}} & \note{5.65/-} & \note{25.70/-} & \note{56.25/-} & \note{66.0/63.64} & \note{70.33/70.69} & \note{72.43/73.32} & \note{73.54/\textbf{74.23}}\\
                SNM \cite{wang2022signed} &                         &73.18 & - & - & - & 64.78 & 71.50 & 72.86 & \note{-}\\
                Burst \cite{li2022efficient} &                         &74.27 & - & - & - & 70.61 & 73.32 & 73.99 & \note{74.25}\\


            { QCFS \cite{bu2021optimal} } &                        &74.29 & - & - & 50.97&68.47 & 72.85 & 73.97 &\note{74.22}\\
            { OPI \cite{bu2022optimized} } &                          &74.85& - & 6.25 &36.02 & 64.70 & 72.47 & 74.24 & \note{74.62}\\

            \midrule \textbf{Ours}                    &       VGG-16 & 74.88 & \textbf{59.95} & \textbf{62.51} & \textbf{70.13} & \textbf{73.44} & \textbf{74.68} & \textbf{74.93} & \note{74.94}\\
\bottomrule
    \end{tabular}
    }
    \end{table*}
\subsection{Tradeoff between Accuracy and Inference Delay}
We compare the proposed method with 
the state-of-the-art ANN-SNN conversion
methods
  on CIFAR-10, CIFAR-100 (Table \ref{tb.acc_vs_delay})\note{,} and ImageNet (Table \ref{tb.imagenet}) datasets. 
\note{  Here, 
in order to better compare the performance with different methods, we provide reproduced experimental results under \noteRA{few time} steps in the format of "\textbf{A/B}".}

In general, the proposed method achieves the best result nearly across all time steps.
For ultra-low latency $T=2$ and $T=4$, our method achieves promising performance improvements of $14.53\%$ and $2.84\%$ with ResNet-18 on \noteRA{the} CIFAR-10 
dataset respectively. 
Notably, SNNs perform better than ANNs when time step increases up to 128 as shown in Table \ref{tb.acc_vs_delay}.
The identified residual potential error (RPE) in this work explains the underlying cause. The spike rate is used in conversion to approximate ANN activation.
 Typically, it is difficult to eliminate the non-deterministic component (i.e., the RPE) of approximation error. Thus, RPE serves as noise when using the firing rate of SNNs to replace the ANN activations (Eq. \ref{eq.sr2sr}).
  The noise becomes the optimizing objective in \note{Stage-II}, and weights and initial potential are finetuned against such noise, so that the RPE brings additional generalization ability to converted SNNs which 
\begin{table}[t]
    \vspace{-6pt}
    \centering
    \caption{ Comparison with direct training with spike-based backpropagation.}
    \vspace{-8pt}

    \begin{threeparttable}
    \label{tb.direct_train} 
    \resizebox*{.96\linewidth}{!}{
    \begin{tabular}{cccc}
    \toprule 
     \textbf{ Method } & \textbf{ Architecture } & \textbf{ Time Steps } & \textbf{ Accuracy (\%) } \\
    \midrule 
    \multicolumn{4}{c}{\textbf{CIFAR-10}}\\
    \midrule
   \noteRA{BRP \cite{zhang2022BRP} $\dagger$}& \noteRA{2D-CSNN} & \noteRA{-} & \noteRA{57.08} \\
      STBP \cite{wu2019direct}  &  { CIFARNet } & 12 & 90.53 \\
       Hybrid \cite{rathi2020enabling} &  { ResNet-20 } & 250 & 92.22 \\
       \noteRA{DIET-SNN \cite{rathi2020diet}} & \noteRA{ResNet-20} & \noteRA{5} & \noteRA{91.78}\\
       TSSL \cite{zhang2020temporal} &  { CIFARNet } & 5 & 91.41 \\
       STBP-tdBN \cite{zheng2021going}  &  { CIFARNet } & 4 & 92.92 \\
\cmidrule{1-4}
    \multirow{2}{*}{ \textbf{Ours} } & ResNet-20 & \textbf{4} &  \textbf{93.27} \\
     & VGG-16 & \textbf{4} &  \textbf{94.06} \\
     \midrule
     \multicolumn{4}{c}{\textbf{CIFAR-100}}\\
     \midrule
       Hybrid \cite{rathi2020enabling}  &  { VGG-11} & 125 &67.87 \\
       \noteRA{DIET-SNN} \cite{rathi2020diet} & \noteRA{VGG-16} & \noteRA{5} & \noteRA{69.67}\\

    STBP-tdBN \cite{zheng2021going}  &  { ResNet-19} & 4 &70.86 \\\cmidrule{1-4}

     \multirow{2}{*}{ \textbf{Ours} } & ResNet-18 & \textbf{4} &  \textbf{75.98} \\
     & VGG-16 & \textbf{4} &  {70.08} \\

    \bottomrule

    \end{tabular}
    }
    \begin{tablenotes}
      \item    \noteRA{The symbol $\dagger$ indicates that the method incorporates a biologically plausible pseudo-backpropagation approach, leveraging spike-timing-dependent plasticity (STDP).}
    \end{tablenotes}
\end{threeparttable}
    \vspace{-5pt} 
    \end{table}
\begin{table}[]
    \vspace{-5pt}
    \centering
    \caption{\note{The effect of the number $N$ of calibration samples. The number in parentheses implies the accuracy of the source ANN.}}
    \label{tb.num_cal}
    \resizebox{ \linewidth}{!}{
    \begin{tabular}{cccccccc}
        \toprule
        \multirow{2}{*}{\textbf{\# samples}}  & \multicolumn{3}{c}{\textbf { ResNet-18 (96.41) }}& & \multicolumn{3}{c}{\textbf { VGG-16  (95.73)}} \\
        \cmidrule(){2-4} \cmidrule(){6-8}
    & $T=2$ & $T=4$ & $T=8$& &$T=2$ & $T=4$ & $T=8$\\
    \midrule 

         0  & 76.01 & 66.52 & 84.17& & 88.89 & 92.89 & 92.72 \\



       \note{2} & \note{75.55} & \note{32.20} & \note{74.47} && \note{47.54} & \note{17.25} & \note{84.01} \\
        \note{4} & \note{83.46} & \note{80.26} & \note{90.42} && \note{82.92} & \note{51.22} & \note{91.75} \\

        \note{8} & \note{85.79} & \note{87.67} & \note{93.21} && \note{89.04} & \note{88.81} & \note{93.24} \\

         16 & 87.89 & 90.47 & 94.31& & 88.96 & 91.79 & 94.31 \\
         32 & 88.75 & 91.94 & 94.89 && 90.21 & 93.24 & 94.41 \\
         64 & 89.56 & 92.40 & 94.85 && 90.59 & 93.19 & 94.80 \\
         128 & 89.74 & 92.18 & 95.07 && 90.91 & 93.54 & 95.01 \\
         256 & 89.51 & 92.77 & 95.29 && 90.80 & 94.04 & 95.02 \\
         512 & 89.89 & 92.96 & 95.31 && 91.41 & 94.13 & 95.22 \\
         1024 & 90.49 & 93.23 & 95.35 && 90.98 & 93.60 & 95.04 \\

    \bottomrule
    \end{tabular}
    }
    \vspace{-5pt}
\end{table}
\noteRA{their ANN counterparts} do not have. 
The superiority seems to vanish when $T \ge 8$ compared to the most recent work \cite{bu2021optimal}.
However, when considering large-scale datasets like ImageNet,
the proposed method still \noteRA{improves} the state-of-the-art of accuracy-delay tradeoff sustainedly as explicated in Table. \ref{tb.imagenet}.
 For instance, the inference delay is shrunk by $4\times$
 into 4 time steps on ImageNet with VGG16 architecture while 
still obtaining significant performance improvements (8.98\% top-1 at least, 59.95\% vs. 50.97\%) 
compared to the existing literature. 
 Notably, as we do not apply the CollorJitter data augmentation as done in
\cite{li2021free,bu2022optimized}, the result of source ANN on ImageNet
 and the best SNN ($T\ge 256$) 
 is slightly lower than 
the best results reported \cite{li2021free}.
Even though, our method still outperforms previous works and achieves 74.68\% top-1
accuracy under relatively long time steps ($T=64$). \note{A full comparison of data augmentation methods, encoding, and decoding methods with the best-performing competitors is provided in Table \ref{tb.1t}. } 

\note{It is worth noting that different methods use different source ANN activation functions (e.g., clipping function in TCL\cite{ho2021tcl}) and different training methods (e.g., customized auxiliary loss for RNL layer in \cite{ding2021optimal}). Hence, achieving exactly the same ANN performance, even with the same data preprocessor and optimizer, is challenging. To further eliminate the influence of the source ANN, we assess the relative performance losses $\Delta_\text{Acc}=(\text{Acc}_{\text{ANN}}-\text{Acc}_{\text{SNN}})/\text{Acc}_{\text{ANN}}$ at $T=32$, $T=64$, and $T=128$, as demonstrated in Table \ref{tb.relative}. In general, the proposed method significantly reduces the conversion loss and enhances the performance of target SNNs. } Moreover, as shown in Table \ref{tb.direct_train}, our
method is also competitive compared to the state-of-the-art direct training method characterized by temporal credit assignment under only $4$
time steps on both CIFAR datasets.  
All those results illustrate the effectiveness of optimizing threefold errors simultaneously.

\subsection{Effect of Different Sample Numbers}

\begin{table}[t]
    \setlength\tabcolsep{3pt}
    \centering
    \caption{\note{Comparing the energy cost under different time steps on CIFAR-100. }
    \note{The numbers in parentheses represent relative accuracy losses during conversion.}
    }
    \label{tb.energy}
    \resizebox{ \linewidth}{!}{
    \begin{tabular}{@{}c@{}ccccc@{}}

    \toprule
    \textbf{Metrics} & \textbf{Method} &\textbf{ANN} & $T=2$  & $T=4$  & $T=8$   \\
    \midrule
    \multirow{4}{*}{\textbf{Accuracy (\%)}}
    &  \note{OPI} \cite{bu2022optimized} &\note{79.43} & \note{33.07(73.49)}  & \note{47.16(40.63)}  & \note{65.50(17.54)} \\
    &  \note{CAP} \cite{li2021free} &\note{78.89} & \note{38.37(51.36)}  & \note{53.96(31.60)}  & \note{64.87(17.77)} \\
    &  \note{QCFS\cite{bu2021optimal}} &\note{77.75} & \note{57.05(26.62)}  & \note{67.28(13.47)}  & \note{73.57(5.38)} \\
    \cmidrule{2-6}
    &\textbf{Ours}  &79.22 & 73.20(7.60)  & 75.98(4.09)  & 78.84(0.48) \\
    \midrule \multirow{4}{*}{\textbf{GSOP ($\times$ 1e-3)}} & \note{OPI \cite{bu2022optimized}}&  \note{480.2}  & \note{83.4$\pm$4.0} & \note{159.0$\pm$8.3} & \note{309.4$\pm$17.5} \\
    &  \note{CAP \cite{li2021free}} &\note{480.2} & \note{159.1$\pm$4.5}  & \note{236.5$\pm$9.1}  & \note{368.0$\pm$23.2} \\
    & \note{QCFS\cite{bu2021optimal}}  & \note{480.2} & \note{138.7$\pm$6.2} & \note{266.5$\pm$14.1} & \note{522.1$\pm$30.3} \\
    \cmidrule{2-6}
    & \textbf{Ours}& 480.2 & 156.1$\pm$6.3 & 238.9$\pm$10.0 & 433.7$\pm$17.2 \\
    \midrule
     \multirow{4}{*}{\textbf{ Energy (${\mu \mathrm{J}}$)}} 
     & \note{OPI \cite{bu2022optimized}} & \note{2208.7} & \note{75.1$\pm$3.6} & \note{143.1$\pm$7.4} & \note{278.5$\pm$15.7}  \\
     &  \note{CAP \cite{li2021free}} &\note{2208.7} & \note{143.2$\pm$4.0}  & \note{212.9$\pm$8.2} & \note{331.2$\pm$20.9} \\
     &\note{QCFS\cite{bu2021optimal}}  & \note{2208.7} & \note{124.8$\pm$5.6}& \note{239.8$\pm$12.7} & \note{469.9$\pm$27.2}  \\
     \cmidrule{2-6}
     &\textbf{Ours} & 2208.7 & 140.5$\pm$5.7 & 215.0$\pm$ 9.0 & 356.0$\pm$21.7  \\
    \midrule 
    \multirow{4}{*}{ \textbf{ Energy Ratio(\%) }}
     & \note{OPI} \cite{bu2022optimized} & \note{100} & \note{3.40$\pm$0.16} & \note{6.48$\pm$0.34} & \note{12.61$\pm$0.71}  \\
     &  \note{CAP \cite{li2021free}} &\note{100} & \note{6.48$\pm$0.18}  & \note{9.64$\pm$0.37} & \note{15.00$\pm$0.95} \\
     & \note{QCFS\cite{bu2021optimal}} & \note{100} & \note{5.65$\pm$0.25} & \note{10.86$\pm$0.57} & \note{21.27$\pm$1.23}  \\
     \cmidrule{2-6}
     &\textbf{Ours} & 100 & 6.36$\pm$0.26 & 9.73$\pm$0.41 & 16.11$\pm$0.98  \\
    \bottomrule
    \end{tabular}
        }
\end{table}

\note{As shown in Table \ref{tb.num_cal}, we investigate the impact of calibration sample numbers on CIFAR-10
 using ResNet-34 and VGG-16 architectures. The initial line shows the baseline results without calibration. Interestingly, a small number of calibration samples \noteRA{yield} significant enhancements. 
 For instance, calibration on just 32 samples elevates the performance of spiking ResNet-18 by 25.42\% 
 when executed under 4 time steps, demonstrating the sample efficiency of the proposed calibration approach,
  particularly for low-latency conversion.
   It is worth mentioning that inadequate sample numbers may cause a certain decline in performance due to the randomness of BPTT calibration. 
   In our experience, using 64 calibration samples leads to considerable performance improvement under low inference steps.
    During actual deployment, an appropriate calibration number can be obtained further by analyzing the regression curve of performance gains with respect to calibration samples.}

\subsection{Energy-Efficiency and Sparsity}
HereIn this section, we count the total synaptic operations (SOP)
to estimate the computation overhead of SNN compared to
their ANN counterparts \cite{merolla2014million}.
 Especially, synaptic operation
by accumulation (AC) in SNNs is variable with spike
sparsity while synaptic operation by multiply and accumulation
(MAC) in ANNs is constant given a particular network
structure.  Here, we measure
32-bit float-point AC and MAC by $\alpha_{ \text{AC}} = 0.9 \ pJ$ and
$\alpha_{ \text{MAC}} = 4.6 \ pJ$ per operation individually as done in \cite{han2015learning}.
Notably, the estimation is extremely conservative and the energy consumption of SNNs on specified hardware design \note{can} be reduced by 12$\times$ to 77 fJ/SOP \cite{qiao2015reconfigurable}.

From Table \ref{tb.energy}, SNN is more SOP-efficient with
sparse spike communications than the ANN in the same
ResNet-18 architecture.
The energy efficiency
is more significant considering the operation degrading from
MAC to AC. For instance, the conversion is nearly lossless (0.38\% accuracy loss)
when $T = 8$ with only $16\%$ energy consumption compared to the ANN counterpart.
For edge devices with stringent power requirements,
the energy consumption \noteRA{is} downgraded to
140 $\mu J$ further ( $15.7\times$ energy saving with 73\%+ performance on
CIFAR-100). \note{We further compare the energy consumption with the state-of-the-art methods \cite{bu2021optimal,bu2022optimized,li2021free}.  In this analysis, we ensured that different methods were reproduced under the same conditions, and we used the same set of 1024 samples to estimate the power consumption.  As shown in Table \ref{tb.energy}, although our approach consumes slightly more energy compared to OPI \cite{bu2022optimized}, it is justified by the significant performance gains achieved at \noteRA{few time} steps. Specifically, our method reduces the relative performance loss ($\Delta_\text{Acc}$) by 65.89\% under 2 time steps}.
\subsection{\note{Extensibility for Complex Structures}}
\begin{table}[t]
    \centering
    \caption{\note{Scaling the proposed method to ResNeXt29-2x64d and MobileNetV2 architectures.} }
    \label{tb.complex}
    \note{
    \resizebox{ \linewidth}{!}{
    \begin{tabular}{crcccc}
    \toprule
    \multirow{2}{*}{\textbf{Method}} & \multicolumn{2}{c}{\textbf{ResNeXt29-2x64d (96.55)}} &  & \multicolumn{2}{c}{\textbf{MobileNetV2 (95.29)}}  \\
     \cmidrule{2-3} \cmidrule{5-6}
                                     & $T=8$             & $T=16$             &  & $T=16$           & $T=32$          \\ \midrule
    Baseline & 27.60  & 49.74   &  & 15.59  & 29.36 \\
    \note{Stage}-I    & 79.17  & 92.43  &  & 26.16    & 54.17  \\
    \note{Stage}-II  & 62.96  & 83.92  &  & 60.42   & 88.63                  \\
    \textbf{\note{Stage}-I + \note{Stage}-II}    & \textbf{90.02}   & \textbf{94.33} &  & \textbf{75.64}  & \textbf{91.49}   \\ \bottomrule
    \end{tabular}
    }
    }
    \vspace{-5pt}
    \end{table}

    \begin{table}[t]
        \centering
        \caption{\note{Scaling the proposed method to Leaky ReLU.} }
        \label{tb.lrelu}
        \note{
        \resizebox{ .9\linewidth}{!}{
        \begin{tabular}{crcccc}
        \toprule
        \multirow{2}{*}{\textbf{Method}} & \multicolumn{2}{c}{\textbf{VGG-16 (95.62)}} &  & \multicolumn{2}{c}{\textbf{ResNet-18 (96.81)}}  \\
         \cmidrule{2-3} \cmidrule{5-6}
                                         & $T=4$             & $T=8$             &  & $T=4$           & $T=8$          \\ \midrule
        Baseline & 47.77  & 71.80   &  & 71.46  & 87.49 \\
        \note{Stage}-I    & 78.83  & 82.47  &  & 88.27    &91.55  \\
        \note{Stage}-II  & 86.02  & 91.48  &  & 85.44   & 92.52                  \\
        \textbf{\note{Stage}-I + \note{Stage}-II}    & \textbf{91.76}   & \textbf{93.38} &  & \textbf{91.23}  & \textbf{93.93}  \\ \bottomrule
        \end{tabular}
        }
        }
        \end{table}
\note{
To demonstrate that our method can be applied to more complex models, we conduct ablation studies on the complex ResNeXt29-2x64d \cite{xie2017aggregated} and MobileNetV2 \cite{sandler2018mobilenetv2} structure.
As shown in the \ref{tb.complex}, for both advanced models, the proposed two-stage approach effectively improves the performance of the converted SNN at each stage and operates in a complementary manner. This further confirms the efficiency and adaptability of our technique in handling complex network architectures.
}

Moreover, the applicability of the proposed method extends beyond ReLU-based ANNs to encompass other activation functions, including Leaky ReLU. To achieve this, we follow similar derivations as presented in Section \ref{sc.ea}. Specifically, we utilize spiking neurons with asymmetric thresholds, as described in \cite{yu2021constructing}, and update the QC function, denoted by Eq. \ref{eq.qc}, as follows:
\noteRA{
\begin{equation}
    g_{\theta, T}(x)=\operatorname{clip}\left(\frac{1}{\theta}\lfloor \text{LeakyReLU}(x)\rfloor_{\theta / T}, -1,1\right)
    \label{eq.lqc}
    \end{equation}
}
    \noindent The results of this adaptation are showcased in Table \ref{tb.lrelu}, where we observe a significant improvement in the performance of the converted SNN by stages. \noteRA{Additionally, the latency is dramatically reduced to 8 time steps, marking a substantial advancement compared to the typical state-of-the-art method \cite{yu2021constructing} for conversion with the LeakyReLU activation function, which reports latencies in the hundreds of time steps.}

\begin{table*}[htb]
    \centering
    \caption{Improving the performance of spike-based object detectors through the proposed method. }
    \label{tb.map}
    \resizebox*{.82\linewidth}{!}{
    \begin{tabular}{c|c|ccc|ccc}
    \textbf{Methods}            & \textbf{Time Step}   & \textbf{AP@0.50:0.95}                    & \textbf{AP@0.50}                       & \textbf{AP@0.75}                       & \textbf{AP}$_S$                                                                     & \textbf{AP}$_M$                          & \textbf{AP}$_L$                          \\ \hline
    {Spiking YOLO} \cite{kim2020spiking}      &                      & 0.2                                  & 0.7                                  & 0.1                                  & 0.2                                                                              & 0.5                                  & 0.1                                  \\
    \textbf{\note{Stage-I w/out {Stage}-II}}      &                      & 7.3                                  & 19.7                                 & 3.1                                  & 3.6                                                                              & 6.8                                  & 8.7                                  \\
    \textbf{\note{Stage-I + {Stage}-II}} & \multirow{-3}{*}{20} & 28.5                                 & 64.1                                 & 19.5                                 & 6.2                                                                              & 16.2                                 & 33.5                                 \\ \hline
    {Spiking YOLO} \cite{kim2020spiking}      &                      & 7.9                                  & 22.3                                 & 3.1                                  & 2.9                                                                              & 5.2                                  & 10.3                                 \\
    \textbf{\note{Stage-I w/out {Stage}-II}}      &                      & 20.1                                 & 48.5                                 & 11.6                                 & 4.6                                                                              & 13.0                                 & 24.4                                 \\ 
    \textbf{\note{Stage-I + \note{Stage}-II}} & \multirow{-3}{*}{40} & 30.7                                 & 66.4                                 & 22.5                                 & 7.2 & 17.8                                 & 36.1                                 \\ \hline
    {Spiking YOLO}  \cite{kim2020spiking}    &      2000                & 26.9                                 & 59.9                                & 19.1                                  & 6.7                                                                              & 15.5                                  & 31.6                                 \\
    {Spiking YOLO}  \cite{kim2020spiking}    &      60                & 16.5                                 & 41.6                                 & 8.2                                  & 5.3                                                                              & 9.2                                  & 20.1                                 \\
    \textbf{\note{Stage-I w/out \note{Stage}-II}}      &         60             & 25.7                                 & 58.7                                 & 16.6                                 & 5.7                                                                              & 15.6                                 & 30.6                                 \\
    \textbf{\note{Stage-I + \note{Stage}-II}} & 60 & {{31.2}} & {{66.8}} & { {23.2}} &  {7.1}                                              & { {17.9}} & {{36.7}}
    \end{tabular}
    }
    \end{table*}

\subsection{Improvements on Object Detection}
\begin{figure}[]
    \centering
\resizebox*{\linewidth}{!}{
    \includegraphics*[width=\linewidth]{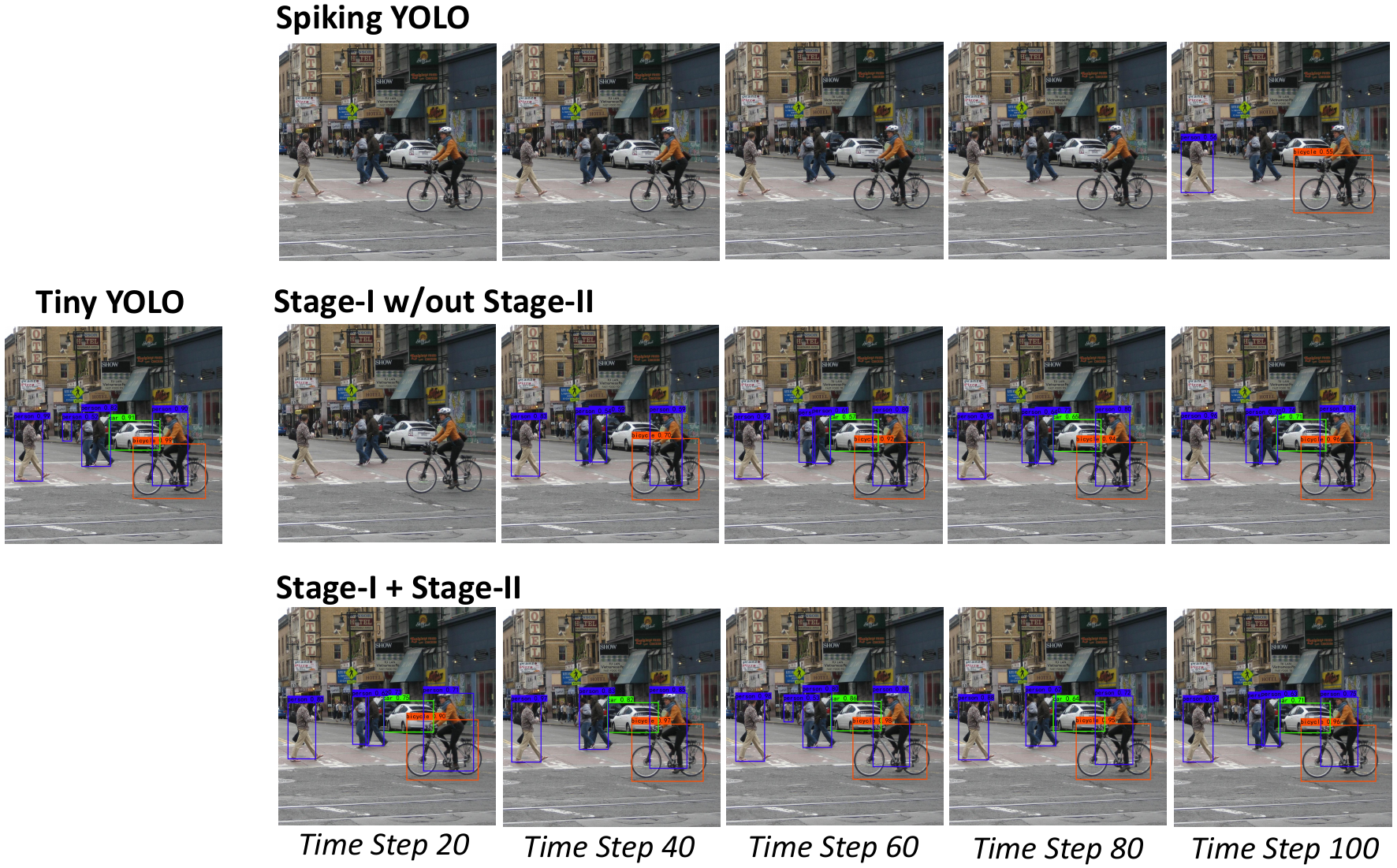}
}
\vspace{-8pt}
    \note{\caption{\note{The comparison of detection results on a single sample with increasing time steps. }}
    \label{fig.detect}
    }
    \vspace{-10pt}
\end{figure}
 In this section, we investigate the effect of the proposed method on a more challenging detection task that involves the coordinate regression of bounding boxes 
 and so requires higher numerical precision than object \note{recognition} to approximate the accurate coordinates.
 For a fair comparison, we implement the well-known spike-based object detector (Spiking YOLO \cite{kim2020spiking}) in the same architecture which achieves 
 better performance than that reported in the original paper.
 As depicted in Table \ref{tb.map}, the proposed method improves continuously the performance of spike-based detection across different object scales and time steps on the PASCAL VOC dataset.
Meanwhile, the detection delay is reduced significantly by at least 25 times to 40 time steps for nearly lossless conversion while SpikingYOLO typically needs thousands of time steps.
 Moreover, \note{Stage-I + {Stage}-II} under \noteRA{few time} steps (e.g. 20 time steps) outperforms \note{Stage-I w/out {Stage}-II} by a large
 margin, especially on the large and middle object detection (33.5\% vs. 8.7\%, 16.2\% vs. 6.8\%). It further illustrates the effect
 of \note{Stage-II} and the importance of optimizing RPE in addition to QE and CE. 

\begin{figure}[]
    \centering
\resizebox*{0.9\linewidth}{!}{
    \includegraphics*[width=\linewidth]{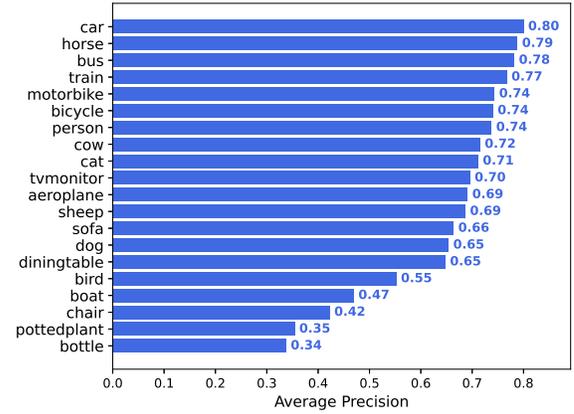}
}
\vspace{-8pt}
    \caption{\noteRA{Average detection precision for each class in 30 time steps.}}
    \label{fig.cls_ap}
\end{figure}

Furthermore, we show the detection results on a sample image with increasing time steps (\note{Fig.} \ref{fig.detect}). 
The far left gives the result of ANN-based Tiny YOLO. The top row displays the detection results of Spiking YOLO \cite{kim2020spiking}. 
In this example, no object can be detected until the time step grows to 100. By optimizing QE and CE, \note{Stage-I} shortens the detection delay to 40 time steps although the numerical precision is still not high.
The \noteRA{delay} is further reduced to 20 time steps and higher detection accuracy is attained by further optimizing RPE in \note{Stage-II}.
We also draw the detection results of each class of objects, as shown in \note{Fig.} \ref{fig.cls_ap}.


\section{Conclusion and Discussion}
\note{In this work, instead of merely optimizing quantization and clipping error
\noteRA{as} in the previous works, \noteRA{this work first explicitly
identifies the errors from residual potential which exhibit a more substantial role in low-latency conversion.}}
\noteRA{
By including an additional finetuning stage to specifically minimize this source of error, 
this study presents a novel {two-stage} converting strategy and obtains high-accuracy and low-latency SNNs.}
The \note{introduced} calibration mechanism considering temporal dynamics in SNNs 
effectively smooths the information loss of knowledge transfer between ANNs and SNNs.
Experiments on large-scale recognition and detection tasks demonstrate our method improves both performance and latency greatly with attractive
power conservation compared to ANNs, which \note{would} facilitate the future application of SNNs on resource-constrained edge devices. 
\noteRA{
Additionally, each module within the pipeline has undergone meticulous testing, including ablation studies, to ascertain its effectiveness and efficiency.
}
For example, attractive performance improvements \noteRA{were} achieved by calibrating with dozens of samples.
We hope that by presenting a fresh perspective on conversion error, this study will help pave the road toward lossless ANN-SNN conversion under extremely low latency.

\note{
    The proposed conversion method has limitations in processing time-domain data with RNN-like and attention-based architectures. In future work, we expect to adapt the concept of temporal calibration to other types of network structures, such as LSTM and Transformer. Additionally, we consider conversion methods based on temporal coding with BPTT calibration, aiming for ultimate energy efficiency and minimized latency. 
    It is important to note that the proposed conversion method is not applicable to online or on-chip learning, as the conversion algorithms generally do not target online or on-chip learning with neuromorphic uses, \noteRA{Instead,} they are widely applicable in efficient inference with lightweight parameter tuning.
Therefore, 
how to realize transfer learning from ANNs to SNNs in a flexible, local, and online way is a worthwhile avenue for future research, especially in the context of deployment on neuromorphic chips.
}
\normalem 

\bibliographystyle{IEEEtran}
\bibliography{ref}

\end{document}